\documentclass{article}

\usepackage[nonatbib,final]{include/neurips_2022}

\usepackage[utf8]{inputenc} %
\usepackage[T1]{fontenc}    %
\usepackage{hyperref}       %
\usepackage{url}            %
\usepackage{booktabs}       %
\usepackage{amsfonts}       %
\usepackage{nicefrac}       %
\usepackage{microtype}      %
\usepackage{xcolor}         %

\usepackage{titletoc}
\usepackage{amsmath}
\usepackage{amsthm}
\usepackage{amssymb}
\usepackage{subfig}
\usepackage{multirow}
\usepackage{array}
\usepackage{multicol}
\usepackage{tabularx}
\usepackage[authoryear]{natbib}

\usepackage[utf8]{inputenc}
\usepackage{booktabs} %
\usepackage{bbm} 
\usepackage{bm}
\usepackage{verbatim}
\usepackage{float}
\usepackage{enumitem}
\usepackage{mathtools}
\usepackage{hhline}
\usepackage[title]{appendix}
\usepackage[nameinlink]{cleveref} %
\usepackage[font=small,labelfont=bf,
   justification=justified,
   format=plain]{caption}
\usepackage{wrapfig,booktabs}
\usepackage{xspace}
\usepackage{cancel}
\usepackage{sidecap}
\DeclarePairedDelimiterX{\infdivx}[2]{(}{)}{%
  #1\;\delimsize\|\;#2%
}

\newcommand{\bnn}{\textsc{bnn}\xspace}
\newcommand{\bnns}{\textsc{bnn}s\xspace}

\newcommand{\fsvi}{\textsc{fsvi}\xspace}
\newcommand{\mfvi}{\textsc{mfvi}\xspace}

\newcommand{\kld}{KL divergence\xspace}

\newcommand{\real}{\mathbb{R}}

\crefname{appsec}{appendix}{appendices}
\Crefname{appsec}{Appendix}{Appendices}

\newtheorem{innercustomproposition}{Proposition}
\newenvironment{customproposition}[1]
  {\renewcommand\theinnercustomproposition{#1}\innercustomproposition}
  {\endinnercustomproposition}

\newcommand{\bX}{\mathbf X}
\newcommand{\by}{\mathbf y}
\newcommand{\bx}{\mathbf x}

\newcommand{\dee}{\,\textrm{d}}

\newcommand{\calF}{\mathcal{F}}

\newcommand{\calN}{\mathcal{N}}

\newcommand{\ftilde}{\smash{\tilde{f}}}

\newcommand{\R}{\mathbb{R}}

\newcommand{\bmu}{{\boldsymbol{\mu}}}
\newcommand{\btheta}{{\boldsymbol{\theta}}}

\newcommand{\bSigma}{\boldsymbol\Sigma}

\newcommand{\closer}[3]{{\kern-#1ex{#2}\kern-#3ex}}

\newcommand{\DKL}{\DD_{\textrm{KL}}\infdivx}

\mathchardef\mhyphen="2D

\DeclareMathOperator{\E}{\mathbb{E}}
\DeclareMathOperator{\Var}{\mathbb{V}}

\newcommand{\edit}[1]{\textcolor{black}{#1}}

\usepackage{algorithm}
\usepackage{algorithmic}
\usepackage{colortbl}

\usepackage{siunitx}
\usepackage{standalone}
\newcommand\defines{\,\dot{=}\,}

\newcommand{\map}{\textsc{map}\xspace}

\newcommand{\swag}{\textsc{swag}\xspace}
\newcommand{\mcd}{\textsc{mc dropout}\xspace}

\newcommand{\vbar}{\,|\,}

\newcommand{\jac}{\mathcal{J}}
\newcommand{\calX}{\mathcal{X}}
\newcommand{\calY}{\mathcal{Y}}
\newcommand{\calD}{\mathcal{D}}
\newcommand{\calB}{\mathcal{B}}
\newcommand{\calQ}{\mathcal{Q}}
\newcommand{\calC}{\mathcal{C}}

\newcommand{\DD}{\mathbb{D}}

\newcommand{\bepsilon}{\boldsymbol{\epsilon}}

\newcommand{\bTheta}{\boldsymbol{\Theta}}

\newcommand{\qtilde}{\smash{\tilde{q}}}
\newcommand{\ptilde}{\smash{\tilde{p}}}
\newcommand{\flin}{\smash{\tilde{f}}}

\newcommand{\cX}{\mathbf{X}_{\calC}}

\newcommand{\pms}[1]{\ensuremath{{\scriptstyle\pm #1}}}

\usepackage[algo2e]{algorithm2e}

\definecolor{mydarkblue}{rgb}{0,0.08,0.45}
\hypersetup{ %
    pdftitle={},
    pdfauthor={},
    pdfsubject={},
    pdfkeywords={},
    pdfborder=0 0 0,
    pdfpagemode=UseNone,
    colorlinks=true,
    linkcolor=mydarkblue,
    citecolor=mydarkblue,
    filecolor=mydarkblue,
    urlcolor=mydarkblue,
    pdfview=FitH
}

\title{Tractable Function-Space Variational Inference in\\Bayesian Neural Networks}

\author{%
  \hspace*{-3pt}Tim G. J. Rudner\thanks{Corresponding author. Email: \href{mailto:tim.rudner@cs.ox.ac.uk}{\texttt{<tim.rudner@cs.ox.ac.uk>}}.} \\
  \hspace*{-3pt}University of Oxford
  \And
  \hspace*{-3pt}Zonghao Chen \\
  \hspace*{-3pt}University College London
  \And
  \hspace*{-3pt}Yee Whye Teh \\
  \hspace*{-3pt}University of Oxford
  \And
  \hspace*{-3pt}Yarin Gal\hspace*{-3pt} \\
  \hspace*{-3pt}University of Oxford\hspace*{-3pt}
}

\begin{document}

\maketitle

\begin{abstract}
Reliable predictive uncertainty estimation plays an important role in enabling the deployment of neural networks to safety-critical settings. A popular approach for estimating the predictive uncertainty of neural networks is to define a prior distribution over the network parameters, infer an approximate posterior distribution, and use it to make stochastic predictions. However, explicit inference over neural network parameters makes it difficult to incorporate meaningful prior information about the data-generating process into the model. In this paper, we pursue an alternative approach. Recognizing that the primary object of interest in most settings is the distribution over functions induced by the posterior distribution over neural network parameters, we frame Bayesian inference in neural networks explicitly as inferring a posterior distribution over functions and propose a scalable function-space variational inference method that allows incorporating prior information and results in reliable predictive uncertainty estimates. We show that the proposed method leads to state-of-the-art uncertainty estimation and predictive performance on a range of prediction tasks and demonstrate that it performs well on a challenging safety-critical medical diagnosis task in which reliable uncertainty estimation is essential.
\end{abstract}

\section{Introduction}
\label{sec:intro}

Machine learning models succeed at an increasingly wide range of narrowly defined tasks~\citep{alexnet,mnih2013playing,alphago,alphafold} but may fail without warning when used on inputs that are meaningfully different from the data they were trained on~\citep{amodei2016open,hendrycks2021unsolved,rudner2021aisafety,rudner2021robustness}.
To deploy machine learning models in safety-critical environments where failures are costly or may endanger human lives, machine learning methods must be reliable and have the ability to `fail gracefully.'
A promising tool for incorporating fail-safe mechanisms into machine learning systems, predictive uncertainty quantification allows machine learning models to express their confidence in the correctness of their predictions.

In this paper, we develop a method for obtaining reliable uncertainty estimates in Bayesian neural networks (\bnns,~\citet{neal1996bayesian}).
While \bnns have promised to combine the advantages of deep learning and Bayesian inference, existing approaches for approximate inference in \bnns fall short of this promise and have been demonstrated to result in approximate posterior predictive distributions that underperform `non-Bayesian' methods both in terms of predictive accuracy and uncertainty quantification---making them of limited use in practice~\citep{ovadia2019uncertainty,foong2019inbetween,farquhar2020radial,Band2021benchmarking}.
A potential reason for this shortcoming is that commonly used parameter-space inference methods make it difficult to define meaningful priors that effectively incorporate information about the data-generating process into inference.

To avoid this limitation, we follow~\citet{sun2019fbnn} and consider a variational objective defined explicitly in terms of distributions over \emph{functions} induced by distributions over parameters.
In contrast to prior works that rely on approximation techniques that prevent such function-space variational objectives to be used with high-dimensional inputs and highly-overparameterized neural networks, we propose a simple estimator of the Kullback-Leibler divergence between distributions over functions that enables us to perform stochastic variational inference.
The proposed estimation procedure allows defining priors that explicitly encourage high predictive uncertainty away from the training data as well as priors that reflect relevant information about the task at hand.

\begin{figure*}[t!]
\centering
    \hspace*{-10pt}
    \subfloat[Predictive Distribution\hspace{-10pt}]{
    \label{fig:illustrative_snelson_small_fsvi}%
        \includegraphics[height=3.03cm, keepaspectratio,trim={0 3pt 0 0},clip]{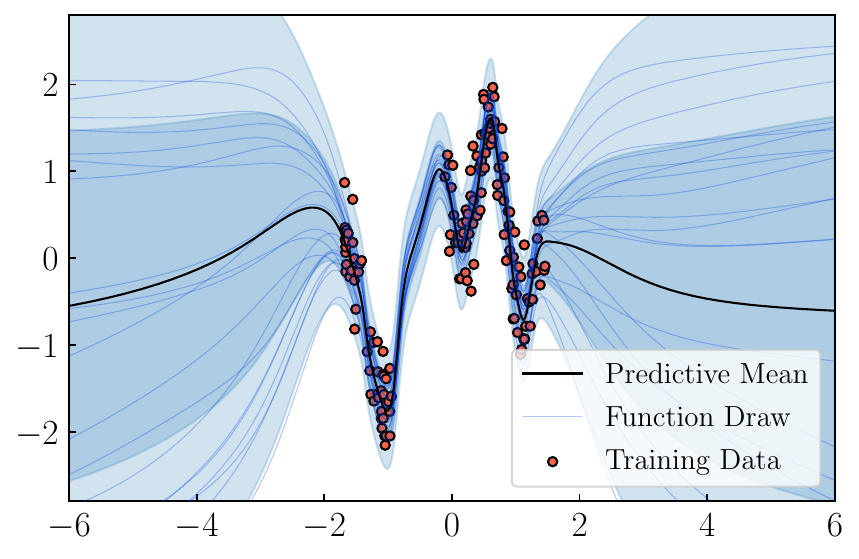}}%
    \subfloat[Predictive Mean~~~~~]{
    \label{fig:illustrative_two_moons_small_mean}%
        \includegraphics[height=3cm]{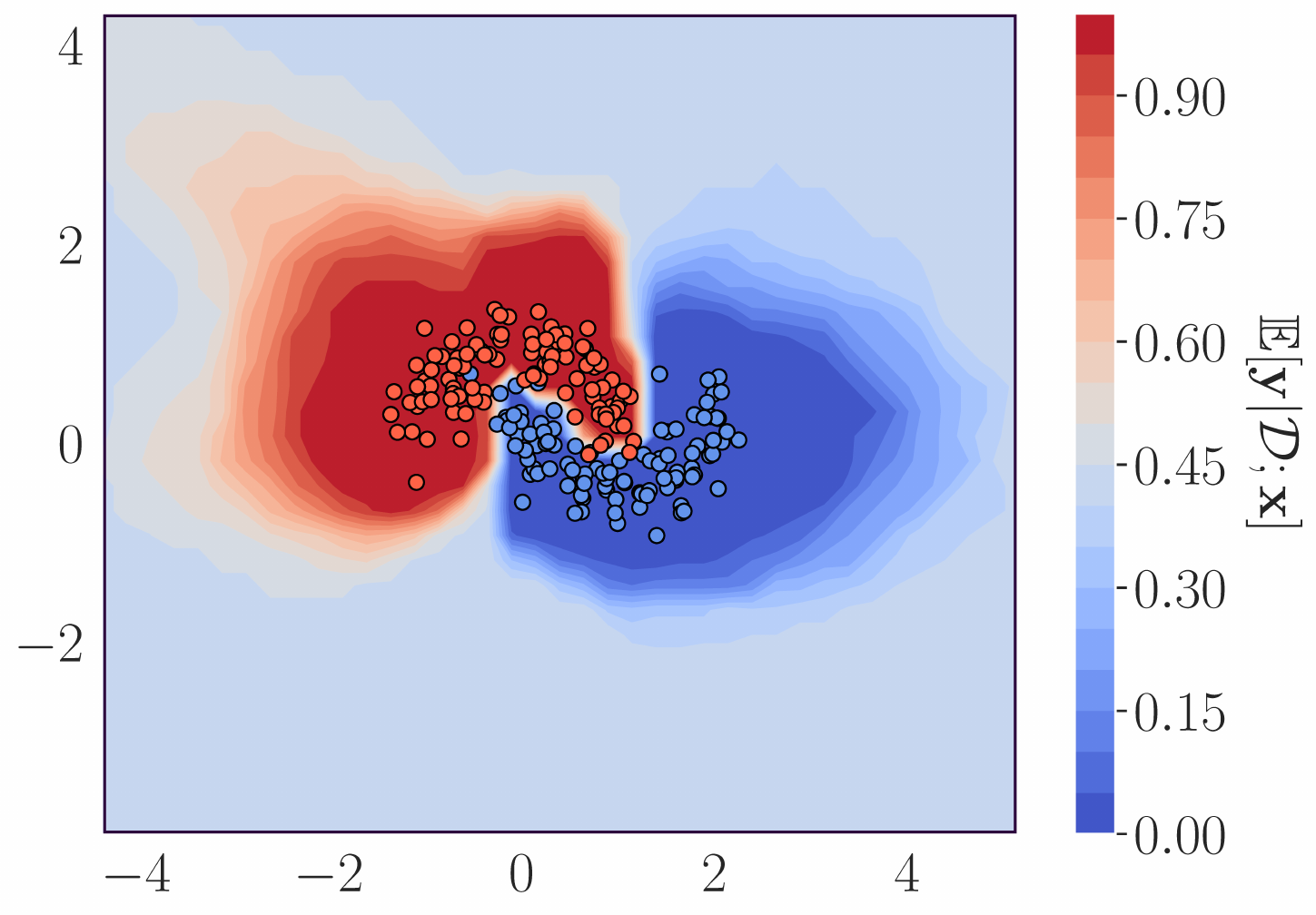}}%
    \hspace*{-5pt}
    \subfloat[Predictive Variance~~~~]{
    \label{fig:illustrative_two_moons_small_var}%
        \includegraphics[height=3cm, keepaspectratio]{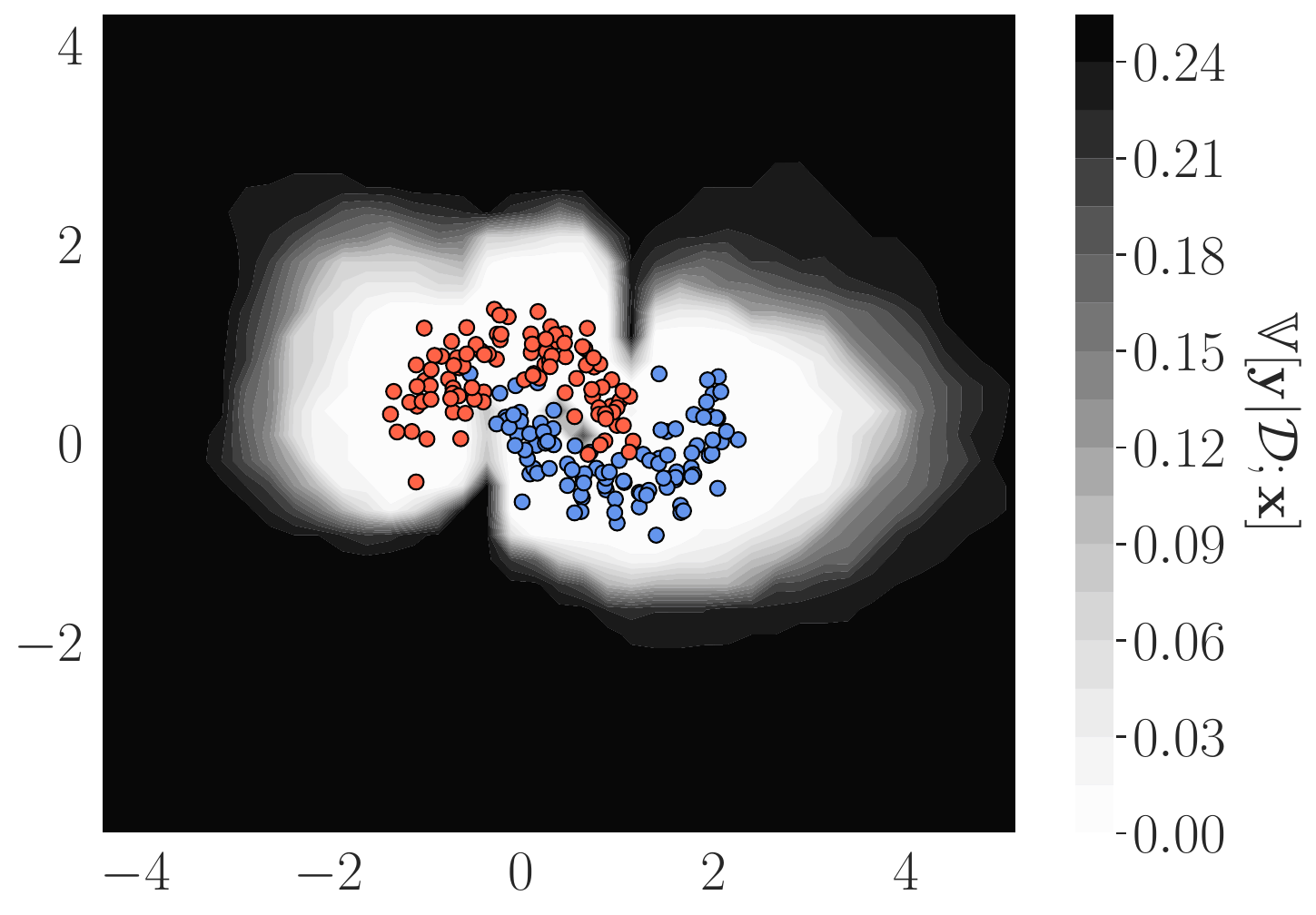}}%
    \caption{
        1D regression on the \textit{Snelson} dataset and binary classification on the \textit{Two Moons} dataset. 
        The plots show the predictive distributions of a \bnns, obtained via function-space variational inference (\fsvi).
        For further illustrative exampled and comparisons to deep ensembles and \bnns learned via parameter-space variational inference, see~\Cref{appsec:empirical_results}.
      }
      \label{fig:illustrative}
    \vspace*{-5pt}
\end{figure*}

We demonstrate that this approach leads to posterior approximations that exhibit significantly improved predictive uncertainty estimates compared to a wide array of state-of-the-art Bayesian and non-Bayesian methods.
\Cref{fig:illustrative} shows examples of predictive distributions obtained via function-space variational inference on low-dimensional, easy-to-visualize datasets.
As can be seen in the figures, the predictive distributions fit the training data well while also exhibiting a high degree of predictive uncertainty in parts of the input space far away from the training data, as desired.

\textbf{Contributions.}$~$
We propose a simple estimation procedure for performing function-space variational inference in \bnns.
The variational method allows for the incorporation of meaningful prior information about the data-generating process into the inference and produces reliable predictive uncertainty estimates.
We perform a thorough empirical evaluation in which we compare the proposed approach to a wide array of competing methods and show that it consistently results in high predictive performance and reliable predictive uncertainty estimates, outperforming other methods in terms of predictive accuracy, robustness to distribution shifts, and uncertainty-based detection of distributionally-shifted data samples.
We evaluate the proposed method on standard benchmarking datasets as well as on a safety-critical medical diagnosis task in which reliable uncertainty estimation is essential.\footnote{Our code can be accessed at \href{https://github.com/timrudner/FSVI}{\underline{\texttt{https://github.com/timrudner/FSVI}}}.}

\section{Preliminaries}
\label{sec:preliminaries}

We consider supervised learning tasks on data \mbox{$\calD \defines \{ (\bx_{n}, \by_{n}) \}_{n=1}^N = (\bX_{\calD}, \by_{\calD})$} with inputs \mbox{$\bx_{n} \in \calX \subseteq \real^D$} and targets \mbox{$\by_{n} \in \calY$}, where $\calY \subseteq \real^Q$ for regression and $\calY \subseteq \{0, 1\}^Q$ for classification tasks.
Bayesian neural networks (\bnns) are stochastic neural networks trained using (approximate) Bayesian inference.
Denoting the parameters of such a stochastic neural network by the multivariate random variable $\bTheta \in \R^P$ and letting the function mapping defined by a neural network architecture be given by \mbox{$f : \calX \times \R^P \rightarrow \R^{Q}$}, we obtain a random function $f(\cdot \,; \bTheta)$.
For a parameter realization $\btheta$, we obtain a corresponding function realization, $f(\cdot \,; \btheta)$.
When evaluated at a finite collection of points $ \bX = \{ \bx_{i} \}_{i=1}^{m} $, $f(\bX ; \bTheta)$ is a multivariate random variable and $f(\bX ; \btheta)$ is a vector.

Letting $p_{\by | f(\bX ; \bTheta)}$ be a likelihood function and $p_{\by | f(\bX ; \bTheta)}(\by_{\calD} \vbar f(\bX_{\calD} ; \btheta))$ be the likelihood of observing the targets $\by_{\calD}$ under the stochastic function $f(\cdot \,; \bTheta)$ evaluated at inputs $\bX_{\calD}$ and letting $p_{\bTheta}$ be a prior distribution over the stochastic network parameters $\bTheta$, we can use Bayes' Theorem to find the posterior distribution, $p_{\bTheta | \calD}$~\citep{mckay1992practical,neal1996bayesian}.
However, since the mapping $f$ is a nonlinear function of the stochastic parameters $\bTheta$, exact inference is analytically intractable.
Variational inference is an approach that seeks to sidestep this intractability by framing posterior inference as a variational optimization problem, where the goal is to find a distribution $q_{\bTheta}$ in a variational family $\calQ_{q_{\bTheta}}$ that solves the variational problem \mbox{$\min_{q_{\bTheta} \in \calQ_{q_{\Theta}}} \DKL{q_{\bTheta}}{p_{\bTheta | \calD}}$}~\citep{wainwright2008vi}.
If $\calQ_{q_{\bTheta}}$ is the family of mean-field Gaussian distributions and the prior distribution over parameters $p_{\bTheta}$ given by a diagonal Gaussian distribution, the resulting variational objective is amenable to stochastic variational inference and can be optimized using gradient-based methods~\citep{hinton1993keeping,graves2011practical,hoffman2013svi,blundell2015mfvi}.

\subsection{A Function-Space Perspective on Variational Inference in Bayesian Neural Networks}
\label{sec:theory}

Instead of seeking to infer an approximate posterior distribution over parameters, we frame variational inference in stochastic neural networks as inferring an approximation to the posterior distribution over \emph{functions} $ p_{f(\cdot \,; \bTheta) | \calD}$ induced by the posterior distribution over parameters $p_{\bTheta | \calD}$, that is,
\begin{align}
\SwapAboveDisplaySkip
    p_{f(\cdot \,; \bTheta) | \calD}(f(\cdot \,; \btheta) \vbar \calD)
    =
    \int_{\R^{P}} p_{\bTheta | \calD}(\btheta' \vbar \calD) \, \delta( f(\cdot \,; \btheta) - f(\cdot \,; \btheta') ) \dee \btheta' ,
\end{align}
where $\delta(\cdot)$ is the Dirac delta function~\citep{wolpert1993fsmap}.
Considering the prior distribution over functions $p_{f(\cdot \,; \bTheta)}$ induced by a prior distribution over parameters $p_{\bTheta}$,
\begin{align}
\SwapAboveDisplaySkip
    p_{f(\cdot \,; \bTheta)}(f(\cdot \,; \btheta))
    =
    \int_{\R^{P}} p_{\bTheta}(\btheta') \, \delta( f(\cdot \,; \btheta) - f(\cdot \,; \btheta') ) \dee \btheta' ,
\end{align}
and the variational distribution over functions $q_{f( \cdot \,; \bTheta)}$ induced by a variational distribution over parameters $q_{\bTheta}$,
\begin{align}
\SwapAboveDisplaySkip
    q_{f( \cdot \,; \bTheta)}(f(\cdot \,; \btheta))
    =
    \int_{\R^{P}} q_{\bTheta}({\btheta'}) \, \delta( f(\cdot \,; \btheta) - f(\cdot \,; \btheta') ) \dee \btheta' ,
\end{align}
we can express the problem of finding a posterior distribution over functions variationally as
\begin{align}
    \min_{q_{\bTheta} \in \calQ_{q_{\bTheta}}} \DD_{\textrm{KL}}( q_{f( \cdot \,; \bTheta)} \,\|\, p_{f(\cdot \,; \bTheta) | \calD} ) ,
\end{align}
which allows us to effectively incorporate meaningful prior information about the underlying data-generating process into training.
As discussed by~\citet{burt2021understanding}, this variational objective is guaranteed to be well-defined for suitably chosen prior distributions over functions.
Specifically, the \kld between two distributions over functions generated from different distributions over parameters applied to the same mapping (e.g., the same neural network architecture) is well-defined (i.e., finite) if the \kld between the distributions over parameters is finite, since, by the strong data processing inequality~\citep{Polyanskiy2017data},
\begin{align}
    \DD_\textrm{KL}(q_{f( \cdot \,; \bTheta)} \,\|\, p_{f(\cdot ; \bTheta)}) \leq \DD_\textrm{KL}(q_{\bTheta} \,\|\, p_{\bTheta}).
\end{align}
As a result, if $\DD_\textrm{KL}(q_{\bTheta} \,\|\, p_{\bTheta}) < \infty$, which is the case for finite-dimensional parameter vectors $\bTheta$ and $q_{\bTheta}$ absolutely continuous with respect to $p_{\bTheta}$, then the function-space \kld is finite and thus well-defined as a variational objective.

Hence, for a likelihood function defined on a finite set of training targets $\by_{\calD}$ and a suitably defined prior distribution over functions, we can express the variational problem above equivalently as the well-defined maximization problem \mbox{$\max_{q_{\bTheta} \in \calQ_{\btheta}} \mathcal{F}(q_{\bTheta})$} with
\begin{align}
\begin{split}
\label{eq:kld_function_space}
    \mathcal{F}(q_{\bTheta})
    &
    \defines
    \mathbb{E}_{q_{f(\bX_{\calD} ; \bTheta)}}[\log p_{\by | f(\bX ; \bTheta)}(\by_{\calD} \vbar f(\bX_{\calD} ; \btheta) )]
    - \DD_\textrm{KL}(q_{f(\cdot ; \bTheta)} \,\|\, p_{f(\cdot ; \bTheta)})
    ,
\end{split}
\end{align}%
where $\DD_\textrm{KL}(q_{f(\cdot ; \bTheta)} \,\|\, p_{f(\cdot ; \bTheta)})$ is also a \kld between distributions over functions.

Unfortunately, evaluating the \kld in~\Cref{eq:kld_function_space} is in general intractable for arbitrary mappings $f$.
To obtain a tractable objective,~\citet{sun2019fbnn} showed that $\DD_\textrm{KL}(q_{f(\cdot ; \bTheta)} \,\|\, p_{f(\cdot ; \bTheta)})$ can be expressed as the supremum of the \kld from $q_{f(\cdot ; \bTheta)}$ to $p_{f(\cdot ; \bTheta)}$ over all \emph{finite} sets of evaluation points, resulting in the objective function
\begin{align}
\begin{split}
\label{eq:kld_fbnn}
    \mathcal{F}(q_{\bTheta})
    =
    \mathbb{E}_{q_{f(\bX_{\calD} ; \bTheta)}}[\log p_{\by | f(\bX ; \bTheta)}(\by_{\calD} \vbar f(\bX_{\calD} ; \btheta ))]
    - \sup_{\bX \in \calX_{\mathbb{N}}} \hspace*{-3pt}\DD_\textrm{KL}(q_{f(\bX ; \bTheta)} \,\|\, p_{f(\bX ; \bTheta)} ) ,
\end{split}
\end{align}%
where $\calX_{\mathbb{N}} \defines \bigcup_{n \in \mathbb{N}} \{ \bX \in \calX_{n} \vbar \calX_{n} \subseteq \mathbb{R}^{n \times D} \}$ is the collection of all finite sets of evaluation points.
However, this objective function is still challenging to optimize in practice:
The supremum cannot be obtained analytically and the \kld term itself is analytically intractable and difficult to estimate in high dimensions---even for a single evaluation point.

In the next section, we will describe an approximation and estimation procedure that allows scaling function-space variational inference to large neural networks and high-dimensional input data.

\section{Deriving a Tractable Function-Space Variational Objective}
\label{sec:fsvi_linearization}

The primary obstacle to computing the objective in~\Cref{eq:kld_function_space} is the \kld from $q_{f(\cdot ; \bTheta)}$ to $p_{f(\cdot ; \bTheta)}$.
There are two reasons why the \kld in~\Cref{eq:kld_fbnn} is intractable:
First, for \bnns or other non-linear models, we do not have access to the probability density functions of the multivariate distributions $q_{f(\bX ; \bTheta)}$ and $p_{f(\bX ; \bTheta)}$;
second, for all but extremely simple input spaces, we are unable to compute the supremum over all possible finite sets of evaluation points.
In the remainder of this section, we outline an approach for obtaining an estimator of a locally accurate approximation to the \kld that allows for scalable gradient-based optimization of~\Cref{eq:kld_fbnn}.

We first approach the problem of computing the \kld between two \bnns evaluated at a finite set of points.
To do so, we first derive tractable approximations to the distributions over functions $q_{f(\bX ; \bTheta)}$ and $p_{f(\bX ; \bTheta)}$
Next, we show that under these approximations, we are able to obtain a closed-form approximation to the \kld and describe a simple Monte Carlo estimator of the supremum in the function-space \kld.

\subsection{Approximating Distributions over Functions via Local Linearization}
\label{subsec:linearizations}

To obtain an approximation to the probability distributions of $q_{f(\bX ; \bTheta)}$ and $p_{f(\bX ; \bTheta)}$, we use a first-order Taylor expansion of the \emph{mapping} $f$ about the mean parameters of $q_{\bTheta}$ and $p_{\bTheta}$, respectively, and derive the induced distributions under the linearized mapping.

\label{prop:linearized_dist}
For a stochastic function $f(\cdot \,; \bTheta)$ defined in terms of stochastic parameters \mbox{$\bTheta$} distributed according to distribution $g_{\bTheta}$ with $\mathbf{m} \defines \E_{g_{\bTheta}}[\bTheta]$ and $\mathbf{S} \defines \text{Cov}_{g_{\bTheta}}[\bTheta]$, we denote the linearization of the stochastic function $f(\cdot \,; \bTheta)$ about $\mathbf{m}$ by
\begin{align}
\label{eq:linearization}
    f(\cdot \,; \bTheta) \approx \flin(\cdot \,; \mathbf{m}, \bTheta) \defines f(\cdot \,; \mathbf{m}) + \jac(\cdot \,; \mathbf{m})(\bTheta - \mathbf{m}),
\end{align}
where $\jac(\cdot \,; \mathbf{m}) \defines (\partial f(\cdot \,; \bTheta) / \partial \bTheta)|_{\bTheta = \mathbf{m}}$ is the Jacobian of $f(\cdot \,; \bTheta)$ evaluated at $\bTheta = \mathbf{m}$, and the mean and covariance of the distribution over the linearized mapping $\flin$ at \mbox{$\bX, \bX' \in \calX$} are given by
\begin{align}
    \E[\flin(\bX ; \bTheta)]
    &
    =
    f(\bX; \mathbf{m})
    \\
    \textrm{{Cov}}[\flin(\bX ; \bTheta), \flin(\bX' ; \bTheta)]
    &
    =
    \jac(\bX ; \mathbf{m}) \mathbf{S} \jac(\bX', \mathbf{m})^\top .
\end{align}
For a derivation of this result, see~\Cref{appsec:proofs}.
Since Gaussianity is preserved under affine transformations, if $g_{\bTheta}$ is a multivariate Gaussian distribution with mean $\mathbf{m}$ and diagonal co-variance $\mathbf{S}$, then the distribution $\tilde{g}$ over $\flin(\bX \,; \bTheta)$ is given by
\begin{align}
\label{eq:dist_lin}
    \tilde{g}_{\flin(\bX ; \mathbf{m}, \bTheta)} = \mathcal{N}(f(\bX ; \mathbf{m}), \jac(\bX ; \mathbf{m}) \mathbf{S} \jac(\bX ; \mathbf{m})^\top) .
\end{align}
For stochastic functions parameterized by many millions of parameters, obtaining the covariance of $\tilde{g}_{\flin(\bX ; \bTheta)}$---which requires computing an inner product of two Jacobian matrices---can be computationally expensive.
Instead of computing the distribution over the linearized mapping exactly, we can construct a suitable Monte Carlo estimator.
To do so, we consider a partition of the set of parameters into sets $\alpha$ and $\beta$ (with $|\beta| \ll |\alpha|$) and note that the linearized mapping can then be expressed as
\begin{align}
\label{eq:linearization_decomposition}
    \flin(\cdot \,; \mathbf{m}, \bTheta)
    =
    f(\cdot \,; \mathbf{m}) + \flin_{\alpha}(\cdot \,; \mathbf{m}, \bTheta_{\alpha}) + \jac_{\beta}(\cdot \,; \mathbf{m})(\bTheta_{\beta} - \mathbf{m}_{\beta}) ,
\end{align}
with
\begin{align}
\SwapAboveDisplaySkip
    \flin_{\alpha}(\cdot \,; \mathbf{m}, \bTheta_{\alpha}) \defines \jac_{\alpha}(\cdot \,; \mathbf{m})(\bTheta_{\alpha} - \mathbf{m}_{\alpha}) ,
\end{align}
where $\jac_{\alpha}(\cdot \,; \mathbf{m})$ and $\jac_{\beta}(\cdot \,; \mathbf{m})$ are the columns of the Jacobian matrix corresponding to the sets of parameters $\alpha$ and $\beta$, respectively, and $\bTheta_{\alpha}$ and $\bTheta_{\beta}$ are the corresponding random parameter vectors.
Noting that~\Cref{eq:linearization_decomposition} expresses $\flin$ as a sum of (affine transformations of) random variables, we can use the fact that for independent Gaussian random variables $\bX$ and $\mathbf{Y}$, the distribution $h_{\mathbf{Z}}$ of $\mathbf{Z} = \bX + \mathbf{Y}$ is equal to the convolution of the distributions $h_{\bX}$ and $h_{\mathbf{Y}}$ to obtain an approximation to $\flin$.
In particular, we can show that if $g_{\bTheta}$ is a multivariate Gaussian distribution with \mbox{$\bTheta_{\alpha} \perp \bTheta_{\beta}$}, the distribution $\tilde{g}_{\flin(\bX ; \bTheta)}$ can be approximated by the Monte Carlo estimator
\begin{align}
\label{eq:dist_lin_approx}
    \hat{\tilde{g}}_{\flin(\bX ; \mathbf{m}, \bTheta)}
    =
    \frac{1}{R} \sum\nolimits_{j=1}^{R} \mathcal{N}\Big(f(\bX ; \mathbf{m}) + \flin_{\alpha}(\bX ; \mathbf{m}, \bTheta_{\alpha})^{(j)}, \jac_{\beta}(\bX ; \mathbf{m}) \mathbf{S}_{\beta} {\jac_{\beta}(\bX ; \mathbf{m})}^\top \Big) ,
\end{align}
where $g_{\bTheta_{\beta}} = \calN(\mathbf{m}_{\beta}, \mathbf{S}_{\beta})$ and samples $\flin_{\alpha}(\bX ; \mathbf{m}, \bTheta_{\alpha})^{(j)}$ are obtained by sampling parameters from the distribution \mbox{$g_{\bTheta_{\alpha}} =  \calN(\mathbf{m}_{\alpha}, \mathbf{S}_{\alpha})$}.
For a derivation of this result, see~\Cref{appsec:proofs}.
This estimator is biased for finite $K$ but converges to $\tilde{g}_{\flin(\bX ; \mathbf{m}, \bTheta)}$ as $R \rightarrow \infty$.
Similarly, for finite $R$, the smaller $ [\mathbf{S}_{\alpha}]_{ii} $, the more accurate and less biased the estimator will be.
In our empirical evaluation, we use a single Monte Carlo sample, $R=1$, to preserve Gaussianity and choose $\alpha$ to be the set of parameters in neural network layers \mbox{$1:L-1$} and $\beta$ to be the set of parameters in the final neural network layer.

\subsection{Approximating the Function-Space Kullback-Leibler Divergence}
\label{subsec:kl_approximation}

From~\Cref{subsec:linearizations}, we know that if $q_{\bTheta}$ and $p_{\bTheta}$ are both Gaussian distributions, then the induced distributions under the linearized mapping $\ftilde$ evaluated at a finite set of evaluation points will be Gaussian as well.
This means that for Gaussian variational and prior distributions over $\bTheta$, we can obtain locally accurate approximations to the induced distributions $q_{f(\cdot ; \bTheta)}$ to $p_{f(\cdot ; \bTheta)}$ and use them to approximate the \kld in the variational objective by $\DD_\textrm{KL}(\qtilde_{\ftilde(\bX ; \bTheta)} \,\|\, \ptilde_{\ftilde(\bX ; \bTheta)})$.
Moreover, for an isotropic Gaussian prior and a mean-field Gaussian variational distribution, $\DD_\textrm{KL}(\qtilde_{\ftilde(\bX ; \bTheta)} \,\|\, \ptilde_{\ftilde(\bX ; \bTheta)})$ is a \kld between two multivariate Gaussians and can be obtained analytically.

Using this approximation, we obtain an estimator of the variational objective given by
\begin{align}
\begin{split}
    &\hspace*{-5pt}\tilde{\mathcal{F}}(q_{\bTheta})
    \defines
    \mathbb{E}_{q_{f(\bX_{\calD} ; \bTheta)}}[\log p_{\by | f(\bX ; \bTheta)}(\by_{\calD} \vbar f(\bX_{\calD} ; \btheta) )]
    - \sup_{\bX \in \calX_{\mathbb{N}}}\DD_\textrm{KL}(\qtilde_{\ftilde(\bX ; \bTheta)} \,\|\, \ptilde_{\ftilde(\bX ; \bTheta)})
    ,
\end{split}
\end{align}%
where the arguments of the \kld have been replaced by the (locally accurate) approximations to the variational and prior distributions over functions evaluated at $\bX$, respectively.
Since the stochastic functions $\ftilde(\cdot \,; \bTheta)$ induced by $q_{\bTheta}$ and $p_{\bTheta}$ under the linearized mapping will be closer to the stochastic function under $f$ the smaller the variance of $q_{\bTheta}$ and $p_{\bTheta}$, respectively, the approximation to the \kld will be more accurate the smaller the variance of $q_{\bTheta}$ and $p_{\bTheta}$.

Next, we turn to computing the supremum.
Unlike~\citet{sun2019fbnn}, who consider the supremum as a separate optimization problem, we do not seek to compute the supremum by searching over points $\bX \in \calX_{\mathbb{N}}$ but instead propose to estimate the supremum at every gradient step via a simple finite-sample estimator.
Specifically, letting $I(\bX) \defines \DD_\textrm{KL}(\qtilde_{\ftilde(\bX ; \bTheta)} \,\|\, \ptilde_{\ftilde(\bX ; \bTheta)})$, we estimate $G = \sup_{\bX \in \calX_{\mathbb{N}}} I(\bX)$ using the Monte Carlo estimator
\begin{align}
\SwapAboveDisplaySkip
\label{eq:kl_estimator_max}
    \hat{G}(\calX_{\calC}^{S})
    =
    \max_{ \bX \in \calX_{\calC}^{S}} I(\bX) ,
\end{align}%
where \mbox{$\calX_{\calC}^{S} \defines \{ \cX^{(i)} \}_{i=1}^S$} is a collection of $S$ sets of \textit{context points} \mbox{$\cX^{(i)} \defines \{ \bx^{(j)} \}_{j=1}^{K}$} jointly sampled from a context distribution $p_{\calX_{\calC}}$.
Each context set $\cX^{(i)}$ can be viewed as a single Monte Carlo sample from the input space so that the estimator $\hat{G}(\calX_{\calC}^{S})$ provides an $S$-sample Monte Carlo estimate of the supremum.
While this estimator is crude and only provides a rough approximation to the true supremum, it encourages the variational distribution over functions to match the prior distribution over functions on the sets of context points.
The choice of the context distribution $p_{\calX_{\calC}}$ can be informed by knowledge about the prediction task and should be viewed as a problem-specific modeling choice.
Similarly, the numbers of samples $S$ and $K$ are hyperparameters to be optimized with a validation set.
For details on how $p_{\calX_{\calC}}$ is chosen for the empirical evaluation in~\Cref{sec:experiments}, see~\Cref{appsec:model_details}.

\subsection{Stochastic Estimation of the Approximate Function-Space Variational Objective}

Let $q_{\bTheta}$ be a Gaussian mean-field variational distribution,  let $p_{\bTheta}$ be an isotropic Gaussian prior, let $(\bX_{\calB}, \by_{\calB})$ be a mini-batch of the training data, and reparameterize $\bTheta$ as $\hat{\bTheta}(\bmu, \bSigma, \bepsilon^{(j)}) \defines \bmu + \bSigma \odot \bepsilon^{(j)}$.
Using the estimator $ \hat{G}(\calX_{\calC}^{S})$ defined above and estimating the expected log-likelihood via Monte Carlo sampling, we obtain a Monte Carlo estimator for the function-space variational objective:
\begin{align}
\label{eq:variational_objective_diag}
    &
    \bar{\calF}(\bmu, \bSigma)
    =
    \frac{1}{M} \sum\nolimits_{j=1}^{M} \log p_{\by | f(\bX ; \bTheta)}(\by_{\calB} \vbar f(\bX_{\calB} ; \hat{\bTheta}(\bmu, \bSigma, \bepsilon^{(j)})) )
    - \max_{ \bX \in  \calX_{\calC}^{S} } {\DD_\textrm{KL}(\qtilde_{\ftilde(\bX ; \hat{\bTheta})} \,\|\, \ptilde_{\ftilde(\bX ; \hat{\bTheta})})}
\end{align}
with $\bepsilon^{(j)} \sim \calN(\mathbf{0}, \mathbf{I}_{P})$ and \mbox{$\calX_{\calC}^{S}$} as defined above.
This Monte Carlo estimator is biased due to the linearization and context-set approximations but allows for scalable gradient-based stochastic optimization.

\textbf{Selection of Prior.}$~$
For all experiments that involve uncertainty quantification, we chose a prior distribution over parameters that induces a prior distribution over functions $p_{f(\cdot ; \bTheta)}$ and a prior predictive distribution that exhibits a high degree of predictive uncertainty at evaluation points from regions in input space where $p_{\calX_{\calC}}$ has non-zero support and, under smoothness constraints, on evaluation points in nearby regions.
For settings where prior information is encoded in data---for example, in the form of expert demonstrations of robotic manipulation tasks~\citep{rudner2021pathologies} or in the form of pre-trained networks in continual or transfer learning~\citep{Rudner2022sfsvi}---an empirical prior that reflects this information can be specified.
For further details, see~\Cref{appsec:model_details}.

\textbf{Selection of Context Distribution.}$~$
The distribution $p_{\calX_{\calC}}$ allows us to incorporate information about the data-generating process into training and encourage the variational distribution to match the prior over functions in relevant parts of the input space.
By taking advantage of the abundance of data available in real-world settings, context distributions can be constructed from large datasets like ImageNet~\citep{alexnet}, from small but diverse datasets like CIFAR-100, or by using any set of task-related unlabeled data.
In our experiments, we choose two types of context distributions.
One of the context distributions is constructed from the training data and only contains randomly sampled monochrome images, and one is constructed from a real-world dataset generated from a data distribution related to that of the training data.
For example, when training on FashionMNIST, we use KMNIST as the context distribution, and when training on CIFAR-10, we use CIFAR-100 as the context distribution.
For further details, see~\Cref{appsec:model_details}.

\textbf{Posterior Predictive Distribution.}$~$
After optimizing the variational objective with respect to the parameters of the variational distribution $q_{\bTheta}$, we use the fact that we can obtain function draws by sampling from the distribution over parameters to obtain an approximate posterior predictive distribution
\begin{align}
\begin{split}
    \hspace*{-3pt}q(\by_{\ast} \vbar \bx_{\ast})
    &
    =
    \int p(\by_{\ast} \vbar f(\bx_{\ast} ; \btheta)) \, q_{f(\bx_{\ast} ; \bTheta)} \, \dee f(\bx_{\ast} ; \btheta)
    \\
    &
    \approx
    \frac{1}{M_{\ast}} \sum\nolimits_{j=1}^{M_{\ast}} p(\by_{\ast} \vbar f(\bx_{\ast} ; \bTheta^{(j)}))
    \quad
    \text{with}
    \quad
    \bTheta^{(j)} \sim q_{\bTheta} ,\hspace*{-5pt}
\end{split}
\end{align}%
where $M_{\ast}$ is the number of Monte Carlo samples used to estimate the predictive distribution.

\section{Related Work}
\label{sec:related_work}

There is a growing body of work on function-space approaches to inference in \bnns, deep learning, and applications such as continual learning~\citep{benjamin2018measuring,sun2019fbnn,titsias2020functional,burt2021understanding,pingbo2020continual,ma2021functional,Rudner2022sfsvi}.

\paragraph{Function-Space Inference in Bayesian Neural Networks.}
Previously proposed methods for \fsvi in \bnns are based on approximate gradient estimators and either replace the supremum in~\Cref{eq:kld_fbnn} with an expectation~\citep{sun2019fbnn} or do not define an explicit variational objective~\citep{wang2018function}.
\citet{sun2019fbnn} and~\citet{Carvalho_2020_CVPR} use Gaussian process priors over functions for which the function-space variational inference problem is not well-defined (see~\Cref{sec:theory} and~\citet{burt2021understanding}).
More recent work has attempted to circumvent the intractability of the variational objective in ~\Cref{eq:kld_function_space} by proposing alternative objectives for function-space inference in \bnns~\citep{ma2019variational,ober2020global,ma2021functional}.
\citet{Rudner2022sfsvi} extend the approach presented in~\Cref{sec:fsvi_linearization} to sequential inference problems and apply it to continual learning.

\paragraph{Linear Models.}
\citet{immer2020improving} and~\citet{khan2019ddn2gp} show that approximate \bnn posterior distribution via the Laplace and Generalized-Gauss-Newton approximation corresponds to exact posteriors under linearizations of different models.
Unlike in our approach, they use a Laplace approximation and do not perform variational inference and do not optimize the variance parameters.
Furthermore,~\citet{immer2020improving} and~\citet{khan2019ddn2gp} use a neural network model to obtain a parameter maximum a posteriori estimate, but then use a linearization of the neural network model to compute a posterior predictive distribution.
In contrast, our work only uses the linearization to obtain an estimator of the variational objective but uses the unlinearized model to construct a posterior predictive distribution.

\paragraph{Pathologies of Variational Inference in Bayesian Neural Networks.}
\citet{burt2021understanding} consider the function-space variational objective in~\Cref{eq:kld_function_space} and show that the \kld between \bnns with different networks architectures are not well-defined.
A parallel line of research showed that posterior predictive distributions of shallow \bnns with mean-field variational distributions have a limited ability to represent complex covariance structures in function space~\citep{foong2019inbetween,foong2020expressiveness} but that deep \bnns do not suffer from this limitation~\citep{farquhar2020liberty}.
Our results are consistent with the findings of~\citet{farquhar2020liberty} that mean-field variational distributions are able to represent complex covariance structures in function space.
\section{Empirical Evaluation}
\label{sec:experiments}
\vspace*{-3pt}

In this section, we evaluate \fsvi on high-dimensional classification tasks that were out of reach for function-space variational inference methods proposed in prior works and compare \fsvi to several well-established and state-of-the-art Bayesian deep learning and deterministic uncertainty quantification methods.
We show that \fsvi (sometimes \emph{significantly}) outperforms existing Bayesian and non-Bayesian methods in terms of their in-distribution uncertainty calibration and out-of-distribution predictive uncertainty estimation.
For a details on models, training and validation procedures, and datasets used, see~\Cref{appsec:model_details}.
For a comparison to~\citet{sun2019fbnn} on small-scale regression tasks, see~\Cref{appsec:uci}.

\vspace{-3pt}
\subsection{Predictive Performance, Uncertainty Estimation, and Distribution Shift Detection}
\label{sec:exps_setup}
\vspace*{-3pt}

In this set of experiments, we assess the reliability of the uncertainty estimates generated by \fsvi.
If a \bnn trained via \fsvi is able to perform reliable uncertainty estimation, its predictive uncertainty will be significantly higher on input points that were generated according to a different data-generating distribution than the training data.
For models trained on the FashionMNIST dataset, we use the MNIST and NotMNIST datasets as out-of-distribution evaluation points, while for models trained on the CIFAR-10 dataset, we use the SVHN dataset as out-of-distribution evaluation points.

For models trained on either FashionMNIST or CIFAR-10, we evaluate their in-distribution performance in terms of test accuracy, test log-likelihood, and test calibration.
To evaluate the quality of different models' uncertainty estimates, we compute uncertainty estimates for the pairs FashionMNIST/MNIST, FashionMNIST/NotMNIST, and CIFAR-10/SVHN to and measure for a range of thresholds how well the datasets in each pair can be separated solely based on the uncertainty estimates.
This experiment setup follows prior work by~\citet{van2020uncertainty} and~\citet{immer2020improving}.
We report the area under the receiver operating characteristic (ROC) curve in Tables~\ref{tab:results_fmnist}~and~\ref{tab:results_cifar}.

\setlength{\tabcolsep}{3pt}
\begin{table*}[t!]
    \caption{
        Comparison of in- and out-of-distribution performance metrics on FashionMNIST (mean $\pm$ standard error over ten random seeds).
        The last two columns show the AUROC for binary in- vs. out-of-distribution detection on MNIST (M) and NotMNIST (NM).
        MNIST and NotMNIST are used as out-of-distribution datasets.
        Best overall results for single and ensemble models are printed in boldface with gray shading.
        Results within a $95$\% confidence interval of the best overall result are printed in boldface only.
        All methods use the same four-layer CNN architecture.
        For further details about model architectures and training and evaluation protocols, see~\Cref{appsec:model_details}.
    }
    \vspace{-5pt}
    \centering
    \small
    \begin{tabular}{l c c c c}
    \toprule
    \textbf{Method}
    &  \textbf{Accuracy} $\uparrow$
    &  \textbf{ECE} $\downarrow$  
    &  \textbf{AUROC} M $\uparrow$ & \textbf{AUROC} NM $\uparrow$
    \\
    \midrule
    \map
     & 91.73\pms{0.08}  & 0.037\pms{0.001} & 87.00\pms{0.30} & 74.85\pms{1.31}
    \\
    \mfvi~\citep{blundell2015mfvi}
     & 91.03\pms{0.04} & 0.038\pms{0.001} & 93.10\pms{0.34} & 88.88\pms{0.74}
    \\
    \mfvi (tempered)
     & 91.38\pms{0.05} & 0.058\pms{0.001} & 86.30\pms{0.29} & 80.78\pms{0.68}
    \\
    \mfvi (radial)~\citep{farquhar2020radial}
     & 90.31\pms{0.11} & 0.035\pms{0.001} & 84.40\pms{0.68} & 82.11\pms{1.15}
    \\
    \mcd~\citep{gal2016dropout}
     & 90.55\pms{0.04} & $0.012$\pms{0.001} & 88.46\pms{0.57} & 80.02\pms{1.04}
    \\
    \swag~\citep{maddox2019swag}
     & 92.56\pms{0.05} & 0.043\pms{0.001} & 85.18\pms{0.35} & 80.31\pms{0.30}
    \\
    \textsc{duq}~\citep{van2020uncertainty}
    & $92.40$\pms{0.20} & $-$ & 95.50\pms{0.70} & 94.60\pms{1.80} 
    \\
    \bnn-\textsc{laplace}~\citep{immer2020improving}
    & 92.25\pms{0.10} & $0.012\pms{0.003}$ & 95.55\pms{0.60} & ~~~~~$-$~~~~~
    \\
    \textsc{spg}~\citep{ma2021functional}
     & 91.60\pms{0.14} & ~~~~~$-$~~~~~ & 95.60\pms{6.00} & ~~~~~$-$~~~~~
    \\
    \fsvi ($p_{\bX_{\calC}}$ = random monochrome)
     & $93.13\pms{0.13}$\hspace{3pt} & $0.012\pms{0.002}$ & ${96.23}\pms{0.46}$ & ${95.02}\pms{0.69}$
    \\
    \fsvi ($p_{\bX_{\calC}}$ = KMNIST)
     & \cellcolor[gray]{0.9}$\mathbf{93.48}\pms{0.12}$\hspace{3pt} & \cellcolor[gray]{0.9}$\mathbf{0.010}$\pms{0.001} & \cellcolor[gray]{0.9}$\mathbf{99.80}\pms{0.20}$ & \cellcolor[gray]{0.9}$\mathbf{97.26}\pms{0.23}$
    \\[1pt]
    \cline{1-5}
    \\[-7pt]
    Deep Ensemble
     & 92.49\pms{0.01} & \cellcolor[gray]{0.9}$\mathbf{0.019}$\pms{0.000}\hspace{3pt} & 89.22\pms{0.09} & 83.17\pms{0.91}
    \\
    \fsvi Ensemble ($p_{\bX_{\calC}}$ = random monochrome)
     & \cellcolor[gray]{0.9}$\mathbf{94.44}$\pms{0.07}\hspace{3pt} & 0.020\pms{0.001} & \cellcolor[gray]{0.9}$\mathbf{97.85}$\pms{0.15} & \cellcolor[gray]{0.9}$\mathbf{96.95}\pms{0.20}$
    \\
    \bottomrule
    \end{tabular}
    \label{tab:results_fmnist}
\end{table*}

\textbf{Predictive Performance and Calibration.}$~$
To assess in-distribution predictive performance and calibration, we report the test accuracy, negative log-likelihood (NLL), and expected calibration error (ECE) for models trained on FashionMNIST and CIFAR-10 in Tables~\ref{tab:results_fmnist}~and~\ref{tab:results_cifar}.
On both \mbox{FashionMNIST} and CIFAR-10, \fsvi achieves the lowest NLL and either the best or second-best predictive accuracy and ECE, respectively, across all methods.
Notably, \fsvi significantly outperforms \textsc{spg}~\citep{ma2021functional}, an alternative function-space variational inference method.

\textbf{Predictive Uncertainty under Distribution Shift.}$~$
In Tables~\ref{tab:results_fmnist}~and~\ref{tab:results_cifar}, we report evaluation metrics that elucidate the reliability of different methods' predictive uncertainty under distribution shift.
\fsvi exhibits reliable predictive uncertainty estimates that allow distinguishing between in- and out-of-distribution inputs with high accuracy.
As would be expected, we observe that using context distributions that reflect our knowledge about the data-generating process can significantly improve uncertainty quantification under \fsvi.
For the FashionMNIST experiment, we used the KMNIST dataset, which contains grayscale images of Kuzushiji letters, and for the CIFAR-10 experiment, we used the CIFAR-100 dataset, which contains RGB images of 100 classes.
Both KMNIST and CIFAR-100 differ from the OOD datasets (MNIST and NotMNIST and SVHN, respectively) used to compute OOD-AUROC metrics in Tables~\ref{tab:results_fmnist} and~\ref{tab:results_cifar}, but using them as context distributions significantly increased the ability of \bnns trained via \fsvi to identify distributionally shifted samples.
Since the variational objective encourages matching the prior (which we chose to have high variance) on samples from the context distribution can improve uncertainty estimation in regions of the input space far from the training data.

\setlength{\tabcolsep}{2pt}
\begin{table*}[t!]
    \caption{
        Comparison of in- and out-of-distribution performance metrics on CIFAR-10 (mean $\pm$ standard error over ten random seeds).
        SVHN and corrupted CIFAR-10 (C-CIFAR) are used as an out-of-distribution datasets.
        The penultimate column shows the AUROC for binary in- vs. out-of-distribution detection on SVHN.
        Best overall results for single and ensemble models are printed in boldface with gray shading.
        Results within a $95$\% confidence interval of the best overall result are printed in boldface only.
        All methods use a ResNet-18 architecture.
        For further details about model architectures and training and evaluation protocols, see~\Cref{appsec:model_details}.
    }
    \vspace{-5pt}
    \centering
    \small
    \begin{tabular}{l c c c c}
    \toprule
    \textbf{Method}
    &  \textbf{Accuracy}$\uparrow$
    &  \textbf{ECE}$\downarrow$  
    &  \textbf{OOD-AUROC}$\uparrow$ & \textbf{C-CIFAR Acc}$\uparrow$
    \\
    \midrule
    \\[-9pt]
    \map
     & 93.19\pms{0.11} & 0.043\pms{0.001} & 94.65\pms{0.27} & 78.87\pms{1.39}
    \\
    \mfvi~\citep{blundell2015mfvi}
     & 89.98\pms{0.09} & 0.040\pms{0.001} & 92.14\pms{0.34} & 79.36\pms{1.35}
    \\
    \mfvi (tempered)
     & 90.87\pms{0.11} & 0.048\pms{0.001} & 91.82\pms{0.90} & 79.86\pms{1.32}
    \\
    \mcd~\citep{gal2016dropout}
     & 93.55\pms{0.07} & 0.040\pms{0.001} & 92.44\pms{0.57} & 80.13\pms{1.37}
    \\
    \swag~\citep{maddox2019swag}
     & 93.13\pms{0.14} & 0.067\pms{0.002} & 89.79\pms{0.50} & 76.12\pms{0.51}
    \\
    \textsc{vogn}~\citep{osawa2019practical}
    & 84.27\pms{0.20} & 0.040\pms{0.002} & 87.60\pms{0.20} & ~~~~~$-$~~~~~
    \\
    \textsc{duq}~\citep{van2020uncertainty}
    & \cellcolor[gray]{0.9}$\mathbf{94.10}$\pms{0.20}\hspace{3pt} & $-$ & 92.70\pms{1.30} & ~~~~~$-$~~~~~
    \\
    \textsc{spg}~\citep{ma2021functional}
     & 77.69\pms{0.64} & ~~~~~$-$~~~~~ & 88.30\pms{4.00} & ~~~~~$-$~~~~~
    \\
    \fsvi ($p_{\bX_{\calC}}$ = random monochrome)
     & $93.35$\pms{0.04} & 0.034\pms{0.001} & $94.76$\pms{0.24} & $\mathbf{80.81}$\pms{0.43}
    \\
    \fsvi ($p_{\bX_{\calC}}$ = CIFAR-100)
     & $93.57$\pms{0.04} & \cellcolor[gray]{0.9}$\mathbf{0.026}$\pms{0.001} & \cellcolor[gray]{0.9}$\mathbf{98.07}$\pms{0.10} & \cellcolor[gray]{0.9}$\mathbf{81.20}$\pms{0.42}
    \\[1pt]
    \cline{1-5}
    \\[-7pt]
    Deep Ensemble
     & $95.13$\pms{0.06} & 0.019\pms{0.001} & $\mathbf{98.04}$\pms{0.07} & $\mathbf{81.22}$\pms{0.37}
    \\
    \fsvi Ensemble ($p_{\bX_{\calC}}$ = random monochrome)
     & \cellcolor[gray]{0.9}$\mathbf{95.19}$\pms{0.03}\hspace{3pt} & \cellcolor[gray]{0.9}\textbf{0.013}\pms{0.001} & \cellcolor[gray]{0.9}$\mathbf{99.19}$\pms{0.41} & \cellcolor[gray]{0.9}$\mathbf{81.35}$\pms{0.48}
    \\
    \bottomrule
    \end{tabular}
    \label{tab:results_cifar}
\end{table*}

\vspace*{-3pt}
\subsection{Generalization and Reliability of Predictive Uncertainty under Distribution Shift}
\vspace*{-3pt}

To assess the reliability of predictive models in deep learning,~\citet{ovadia2019uncertainty} propose the following desiderata:
In order for a model to be considered reliable, it ought to (i) exhibit low predictive uncertainty on training data and high predictive uncertainty on out-of-distribution inputs, (ii) generate predictive uncertainty estimates that allow distinguishing in- from out-of-distribution inputs, and (iii) if possible, maintain high predictive accuracy even under distribution shift.
Models that satisfy these desiderata are less likely to make poor, high-confidence predictions and more amenable for use in safety-critical downstream tasks.

\begin{wrapfigure}{r}{0.45\textwidth}
\centering
\vspace*{-26pt}
    \includegraphics[width=0.85\linewidth, trim={5pt 0 5pt 0},clip]{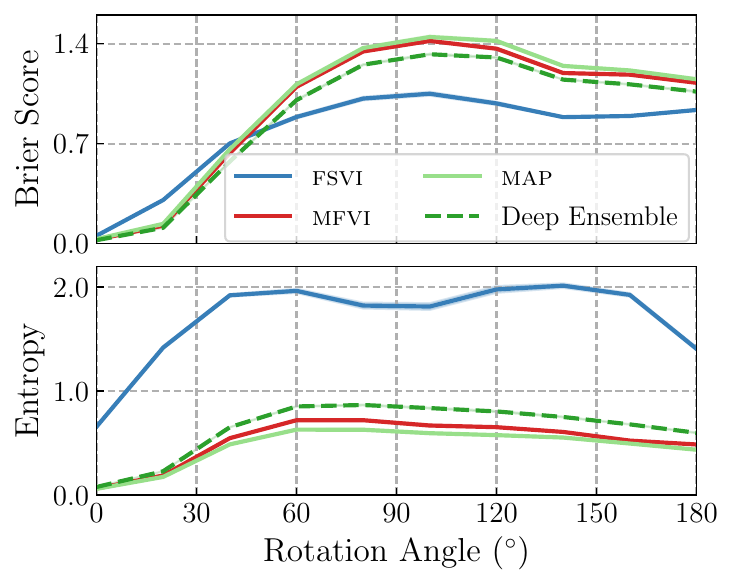}
    \vspace{-5pt}
    \caption{
        Predictive uncertainty and accuracy on rotated MNIST.
        Models with reliable uncertainty estimates would exhibit higher predictive uncertainty the more the digits are rotated.
        Ideally, such models would maintain high predictive accuracy (low Brier score).
    }
    \label{fig:rotated_mnist}
\vspace*{-19pt}
\end{wrapfigure}

To illustrate these desiderata, we follow~\citet{ovadia2019uncertainty} and consider the rotated MNIST task, where a model is trained on MNIST and evaluated on rotated MNIST digits.
The goal is to maintain a high level of predictive accuracy (measured in terms of Brier scores) while exhibiting an increasing level of predictive uncertainty on distribution shifts of increasing magnitude.
\Cref{fig:rotated_mnist} shows Brier scores (lower is better) and predictive entropy estimates (higher means more uncertain) of four different models.
As rotating the MNIST digits gradually shifts the data distributions, we would expect Brier scores to increase (corresponding to worse predictive accuracy) as the rotation angle increases.
A model with reliable predictive entropy estimates would only experience a small decrease under distribution shift while exhibiting a large increase in predictive uncertainty.
As can be seen in the plot, the Brier scores of \fsvi decreases the least, while \fsvi's uncertainty is significantly higher than other models'.
To assess the reliability of different uncertainty quantification methods on a more challenging distribution-shift task, we consider corrupted CIFAR-10 inputs under the second-mildest corruption level used in~\citep{ovadia2019uncertainty} and report our results in~\Cref{tab:results_cifar}.
Consistent with the rotated MNIST results, \fsvi achieves the highest accuracy on the corrupted data.

\vspace*{-3pt}
\subsection{Safety-Critical Uncertainty-Aware Selective Prediction: Diabetic Retinopathy Diagnosis}
\label{sec:diabetic}
\vspace*{-3pt}

\begin{wrapfigure}{r}{0.25\textwidth}
\centering
\vspace*{-15pt}
    \hspace*{-15pt}\includegraphics[width=\linewidth]{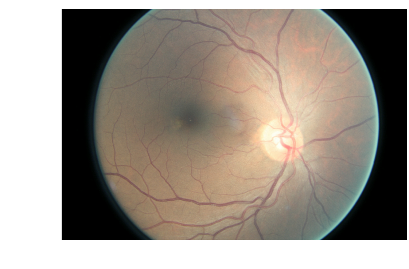}
    \\
    \hspace*{-15pt}\includegraphics[width=\linewidth]{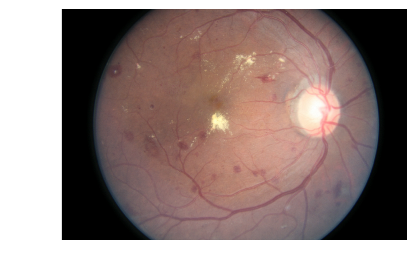}
    \vspace*{-5pt}
    \caption{
        Retina scan examples. \textbf{Top:} healthy. \textbf{Bottom:} unhealthy.
    }
    \label{fig:retina}
\vspace*{-10pt}
\end{wrapfigure}

To evaluate the reliability of the predictive uncertainty of \fsvi in a real-world safety-critical setting, we consider the task of diagnosing diabetic retinopathy (DR), a medical condition that can lead to impaired vision, from retina scans~\citep{leibig2017leveraging,filos2019systematic,Band2021benchmarking}.
We use two publicly available datasets,~\citet{kaggle_2015} and~\citet{APTOS_2019}, each containing RGB images of a human retina graded by a medical expert on the following scale: 0 (no DR), 1 (mild DR), 2 (moderate DR), 3 (severe DR), and 4 (proliferative DR).
The Kaggle dataset was collected from patients in the United States, while the APTOS dataset was collected from patients in India using cheaper but more modern scanning devices.
We follow~\citet{leibig2017leveraging},~\citet{filos2019systematic}, and~\citet{Band2021benchmarking} and binarize all examples from both the EyePACS and APTOS datasets by dividing the classes up into sight-threatening diabetic retinopathy---defined as moderate diabetic retinopathy or worse (classes $\{2,3,4\}$)---and non-sight-threatening diabetic retinopathy---defined as no or mild diabetic retinopathy (classes $\{0,1\}$).
This results in a binary prediction task.

\begin{figure*}[t!]
\centering
\captionsetup[subfigure]{justification=centering}
\includegraphics[width=\linewidth]{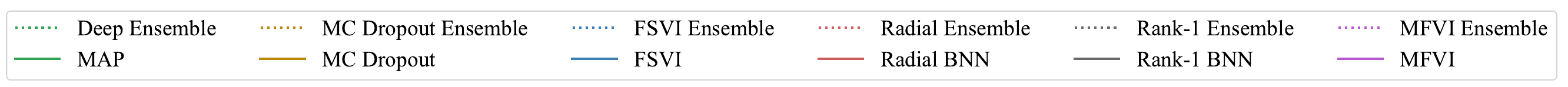}\\[-10pt]
\hspace*{-10pt}\subfloat[
    ROC: In-Domain
]{\includegraphics[width=0.24\linewidth]{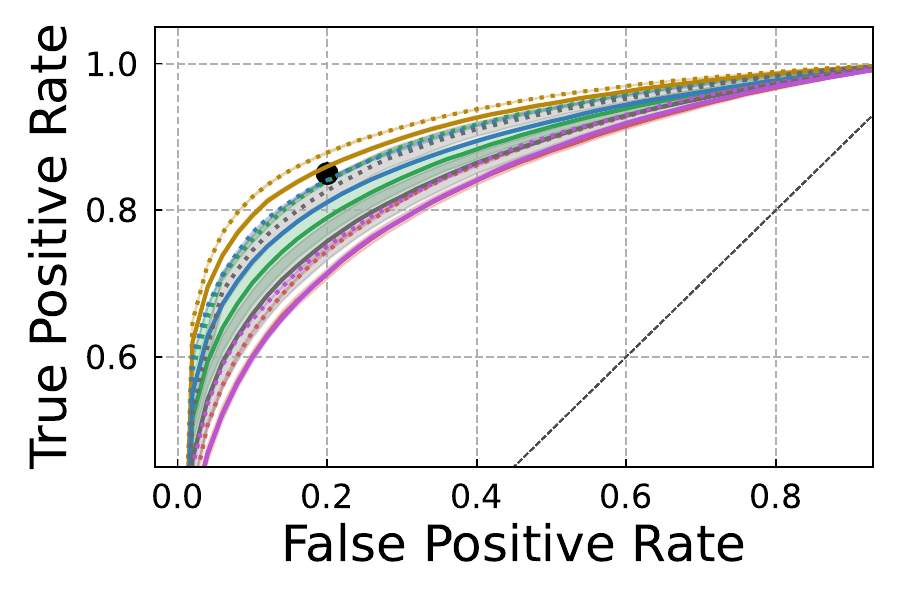}}%
\subfloat[
    ROC: Country Shift
]{\includegraphics[width=0.24\linewidth]{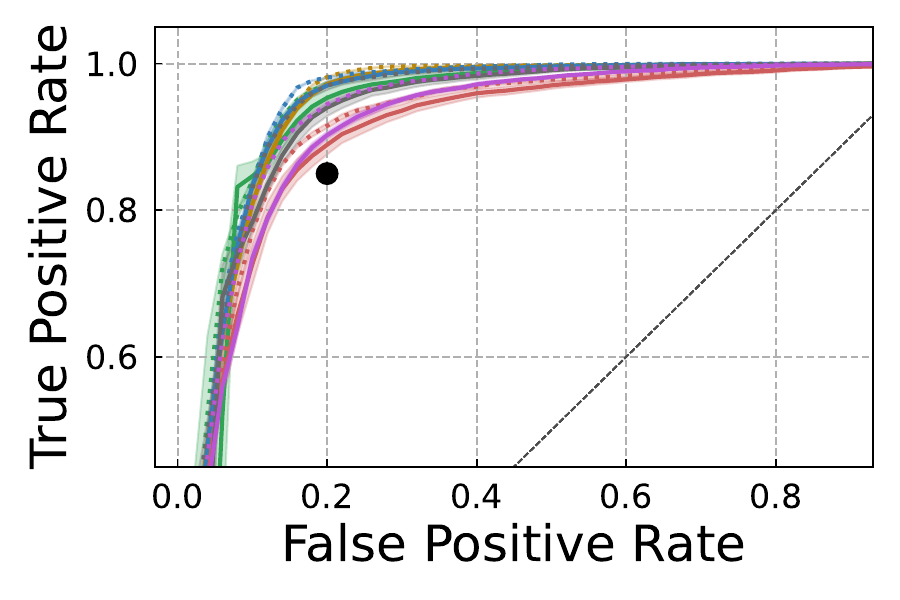}}%
\subfloat[
    Selective Prediction AUROC: In-Domain
]{\includegraphics[width=0.24\linewidth]{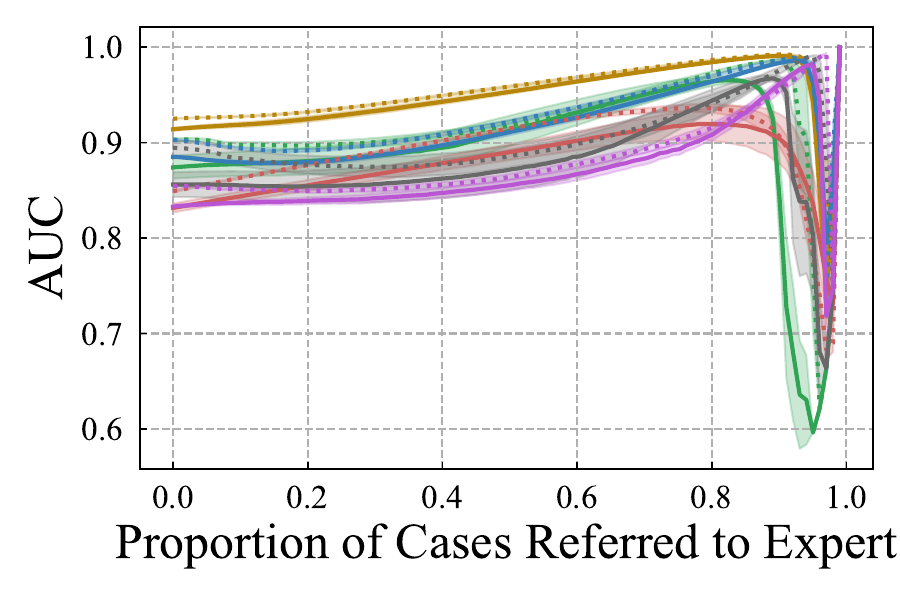}}%
\subfloat[
    Selective Prediction AUROC: Country Shift
]{\includegraphics[width=0.24\linewidth]{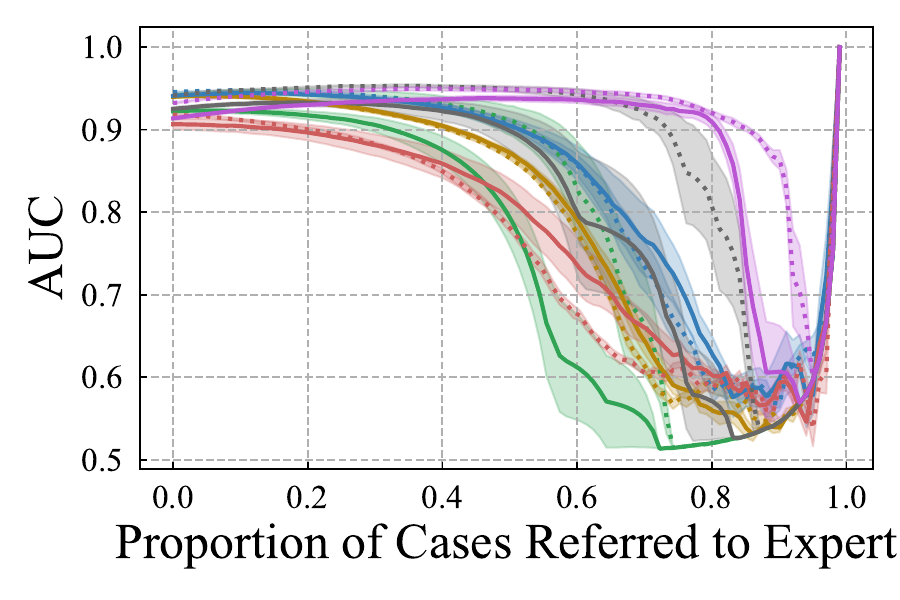}}%
\vspace*{-3pt}
\caption{
    We jointly assess model predictive performance and uncertainty quantification on both in-domain and distributionally shifted data.
    \textbf{Left:} The \textit{receiver operating characteristic curve} (ROC) for in-population diagnosis on the (\textbf{a})~\citet{kaggle_2015} test set and for (\textbf{b}) changing medical equipment and patient populations on the~\citet{APTOS_2019} test set. 
    The dot in 
    \textbf{black}
    denotes the NHS-recommended 85\% sensitivity and 80\% specificity ratios~\citep{widdowson2016management}.
    \textbf{Right:} \emph{Selective prediction} on AUROC in (\textbf{c})~\citet{kaggle_2015} and (\textbf{d})~\citet{APTOS_2019} settings.
    Shading denotes standard error over six random seeds.
    See~\Cref{appsec:retina_tabular} for tabular results.
}
\label{fig:country_shift}
\vspace*{-10pt}
\end{figure*}

To assess the reliability of predictive models when medical training and test data are obtained from different patient populations or collected with the same medical equipment, we follow~\citet{Band2021benchmarking} and use the Kaggle dataset for training and the distributionally shifted APTOS dataset for evaluation.
The results are shown in~\Cref{fig:country_shift}, which plot the ROC curves for the binary prediction problems as well as the area under the ROC curve for an uncertainty aware selective prediction task.
For further details about the uncertainty-aware selective prediction evaluation protocol, see~\Cref{appsec:retina}.
\Cref{fig:country_shift} shows that \fsvi performs well on all four tasks and is only outperformed by \mcd.
For full tabular results, see~\Cref{appsec:retina_tabular}.

\vspace*{-3pt}
\section{Conclusion}
\label{sec:conclusion}
\vspace*{-3pt}

The paper proposed a scalable and effective approach to function-space variational inference in \bnns.
We demonstrated that the proposed estimator of the function-space variational objective can be scaled up to high-dimensional data and large neural network architectures and that \fsvi exhibits consistently reliable in- and out-of-distribution predictive performance on a wide range of datasets when compared to well-established and state-of-the-art uncertainty quantification methods.
We hope that this work will lead to further research into function-space variational inference and the development of more sophisticated data-driven prior distributions over functions.

\clearpage

\section*{Acknowledgements}

We thank Bryn Elesedy, Bobby He, and Andrew Jesson for feedback on an early draft of this paper.
We thank Joost van Amersfoort for helpful discussions about experiment design and implementations.
Tim G. J. Rudner is funded by the Rhodes Trust and the Engineering and Physical Sciences Research Council (EPSRC).
We gratefully acknowledge donations of computing resources by the Alan Turing Institute.

\bibliography{references}
\bibliographystyle{plainnat}

\clearpage

\begin{appendices}

\crefalias{section}{appsec}
\crefalias{subsection}{appsec}
\crefalias{subsubsection}{appsec}

\setcounter{equation}{0}
\renewcommand{\theequation}{\thesection.\arabic{equation}}

\onecolumn

\vspace*{-20pt}
{\hrule height 1mm}

\section*{\LARGE \centering Appendix}
\label{sec:appendix}

\vspace{5pt}
{\hrule height 0.3mm}
\vspace{14pt}

\section*{Table of Contents}
\vspace*{-15pt}
\startcontents[sections]
\printcontents[sections]{l}{1}{\setcounter{tocdepth}{2}}
\vspace*{-10pt}

\clearpage

\section{Proofs \& Derivations}
\label{appsec:proofs}

\subsection{Function-Space Variational Objective}
\label{appsec:function_space_elbo}

This proof follows steps from~\citet{matthews2016kld}.
Consider measures $\hat{P}$ and $P$ both of which define distributions over some function $f$, indexed by an infinite index set $X$.
Let $\calD$ be a dataset and let $\bX_{\calD}$ denote a set of inputs and $\by_{\calD}$ a set of targets.
Consider the measure-theoretic version of Bayes' Theorem~\citep{schervish1995theory}:
\begin{align}
    \frac{d \hat{P}}{d P}(f)=\frac{p_{X}(Y \vbar f)}{p(Y)},
\end{align}
where $p_{X}(Y \vbar f)$ is the likelihood and $p(Y) = \int_{\real^X} p_{X}(Y \vbar f) dP(f)$ is the marginal likelihood.
We assume that the likelihood function is evaluated at a finite subset of the index set $X$.
Denote by $\pi_{C}: \real^{X} \to \real^{C}$ a projection function that takes a function and returns the same function, evaluated at a finite set of points $C$, so we can write
\begin{align}
    \frac{d \hat{P}}{d P}(f)
    =
    \frac{d \hat{P}_{\bX_{\calD}}}{d P_{\bX_{\calD}}}(\pi_{\bX_{\calD}}(f))
    =
    \frac{p(\by_{\calD} \vbar \pi_{\bX_{\calD}}(f))}{p(\by_{\calD})},
\end{align}
and similarly, the marginal likelihood becomes $p(\by_{\calD}) = \int p_{\by | f_{\bX}}(\by_{\calD} \vbar f_{\bX_{\calD}}) \dee P_{\bX_{\calD}}(f_{\bX_{\calD}}) $.
Now, considering the measure-theoretic version of the \kld between an approximating stochastic process $Q$ and a posterior stochastic process $\hat{P}$, we can write
\begin{align}
    \DD_{\textrm{KL}}(Q \,\|\, \hat{P})
    =
    \int \log {\frac{d Q}{d P}(f)} \dee Q(f) - \int \log {\frac{d \hat{P}}{d P}(f)} \dee Q(f),
\end{align}
where $P$ is some prior stochastic process.
Now, we can apply the measure-theoretic Bayes' Theorem to obtain
\begin{align}
    \DD_{\textrm{KL}}(Q \,\|\, \hat{P})
    &=
    \int \log {\frac{d Q}{d P}(f)} \dee Q(f) - \int \log {\frac{d \hat{P}}{d P}(f)} \dee Q(f)
    \\
    &=
    \int \log {\frac{d Q^{\pi}}{d P^{\pi}}(f)} \dee Q^{\pi}(f) - \int \log {\frac{d \hat{P}_{\bX_{\calD}}}{d P_{\bX_{\calD}}}\left(f_{\bX_{\calD}}\right)} \dee Q_{\bX_{\calD}}\left(f_{\bX_{\calD}}\right)
    \\
    &=
    \int \log {\frac{d Q^{\pi}}{d P^{\pi}}(f)} \dee Q^{\pi}(f) - \E_{Q_{\bX_{\calD}}}\left[\log p\left(\by_{\calD} \vbar f_{\bX_{\calD}}\right)\right]-\log p(\by_{\calD}),
\end{align}
where $\frac{d Q^{\pi}}{d P^{\pi}}(f)$ is marginally consistent given the projection $\pi$.
Rearranging, we can get
\begin{align}
    p(\by_{\calD})
    &=
    \E_{Q_{\bX_{\calD}}}\left[\log p_{\by | f_{\bX}}(\by_{\calD} \vbar f_{\bX_{\calD}})\right] - \int \log {\frac{d Q^{\pi}}{d P^{\pi}}(f)} \dee Q^{\pi}(f) + \DD_{\textrm{KL}}(Q^{\pi} \,\|\, \hat{P} )
    \\
    &\geq
    \E_{Q_{\bX_{\calD}}}\left[\log p_{\by | f_{\bX}}(\by_{\calD} \vbar f_{\bX_{\calD}})\right] - \int \log {\frac{d Q^{\pi}}{d P^{\pi}}(f)} \dee Q^{\pi}(f)
    \\
    &=
    \E_{Q_{\bX_{\calD}}}\left[\log p_{\by | f_{\bX}}(\by_{\calD} \vbar f_{\bX_{\calD}})\right] - \DD_{\textrm{KL}}(Q^{\pi} \,\|\, P^{\pi}) .
\end{align}
Finally, this lower bound can equivalently be expressed as
\begin{align}
    p(\by_{\calD})
    &
    \geq
    \E_{Q_{\bX_{\calD}}}\left[\log p_{\by | f_{\bX}}(\by_{\calD} \vbar f_{\bX_{\calD}})\right] - \DD_{\textrm{KL}}(Q_{\bX_{\calD}, \bX_{\backslash \calD}} \,\|\, P_{\bX_{\calD}, \bX_{\backslash \calD}}),
\end{align}
where $\bX_{\backslash \calD}$ is an infinite index set excluding the finite index set $\bX_{\calD}$, that is, $\bX_{\backslash \calD} \cap \bX_{\calD} = \varnothing$, or by Theorem 1 in~\citet{sun2019fbnn}, we can write 
\begin{align}
    p(\by_{\calD})
    &\geq
    \E_{Q_{\bX_{\calD}}}\left[\log p_{\by | f_{\bX}}(\by_{\calD} \vbar f_{\bX_{\calD}})\right] - \sup_{\bX \in \calX_{\mathbb{N}}} \DD_{\textrm{KL}}(Q_{\bX} \,\|\, P_{\bX}),
\end{align}
where $\calX_{\mathbb{N}} \defines \bigcup_{n \in \mathbb{N}} \{ \bX \in \calX_{n} \vbar \calX_{n} \subseteq \mathbb{R}^{n \times D} \}$ is the collection of all finite sets of evaluation points.

\clearpage

\subsection{Distribution under Linearized Function Mapping}
\label{appsec:linearization_proof}

\begin{customproposition}{1}[Distribution under Linearized Mapping]
\label{app-prop:linearized_variational}
For a stochastic function $f(\cdot \,; \bTheta)$ defined in terms of stochastic parameters $\bTheta$ distributed according to distribution $g_{\bTheta}$ with $\mathbf{m} \defines \E_{g_{\bTheta}}[\bTheta]$ and $\mathbf{S} \defines \text{\emph{Cov}}_{g_{\bTheta}}[\bTheta]$, denote the linearization of the stochastic function $f(\cdot \,; \bTheta)$ about $\mathbf{m}$ by
\begin{align*}
    f(\cdot \,; \bTheta) \approx \flin(\cdot \,; \mathbf{m}, \bTheta) \defines f(\cdot \,; \mathbf{m}) + \jac(\cdot \,; \mathbf{m})(\bTheta - \mathbf{m}) ,
    \end{align*}
where $\jac(\cdot \,; \mathbf{m}) \defines (\partial f(\cdot \,; \bTheta) / \partial \bTheta)|_{\bTheta = \mathbf{m}}$ is the Jacobian of $f(\cdot \,; \bTheta)$ evaluated at $\bTheta = \mathbf{m}$.
Then the mean and co-variance of the distribution over the linearized mapping $\flin$ at \mbox{$\bX, \bX' \in \calX$} are given by
\begin{align*}
\SwapAboveDisplaySkip
    \E[\flin(\bX ; \bTheta)]
    &=
    f(\bX; \mathbf{m})
    \\
    \textrm{\emph{Cov}}[\flin(\bX ; \bTheta), \flin(\bX' ; \bTheta)]
    &=
    \jac(\bX ; \mathbf{m}) \mathbf{S} \jac(\bX' ; \mathbf{m})^\top .
\end{align*}
\end{customproposition}
\begin{proof}
We wish to find $\mathbb{E}[\flin(\bX ; \mathbf{m}, \bTheta)]$ and 
\begin{align}
\begin{split}    
    &
    \textrm{Cov}(\flin(\bX ; \mathbf{m}, \bTheta), \flin(\bX'; \mathbf{m}, \btheta))
    \\
    &
    =
    \mathbb{E}[ (\flin(\bX ; \mathbf{m}, \btheta) -  \mathbb{E}[\flin(\bX ; \mathbf{m}, \btheta)])\, (\flin(\bX'; \mathbf{m}, \btheta) -  \mathbb{E}[\flin(\bX'; \mathbf{m}, \btheta)])^\top] .
\end{split}
\end{align}
To see that $\mathbb{E}[\flin(\bX ; \mathbf{m}, \btheta)] = f(\bX ; \mathbf{m})$, note that, by linearity of expectation, we have
\begin{align}
\begin{split}
    \mathbb{E}[\flin(\bX ; \mathbf{m}, \btheta)]
    &
    = \mathbb{E}[ f(\bX ; \mathbf{m}) + \jac(\bX ; \mathbf{m})(\bTheta - \mathbf{m}) ]
    \\
    &
    = f(\bX ; \mathbf{m}) + \jac(\bX ; \mathbf{m})(\mathbb{E}[\bTheta] - \mathbf{m})
    = f(\bX ; \mathbf{m}).
\end{split}
\end{align}
To see that $\textrm{Cov}(\flin(\bX ; \mathbf{m}, \btheta), \flin(\bX'; \mathbf{m}, \btheta)) = \jac(\bX ; \mathbf{m}) \mathbf{S} \jac(\bX' ; \mathbf{m})^\top$, note that in general, for a multivariate random variable $\mathbf{Z}$, $\textrm{Cov}(\mathbf{Z}, \mathbf{Z}) = \mathbb{E}[\mathbf{Z} \mathbf{Z}^\top] + \mathbb{E}[\mathbf{Z}] \mathbb{E}[\mathbf{Z}]^\top$, and hence,
\begin{align}
\begin{split}
    &
    \textrm{Cov}(\flin(\bX ; \mathbf{m}, \bTheta), \flin(\bX'; \mathbf{m}, \bTheta))
    \\
    &
    =
    \mathbb{E}[ \flin(\bX ; \mathbf{m}, \bTheta) \flin(\bX'; \mathbf{m}, \bTheta)^\top ] - \mathbb{E}[\flin(\bX ; \mathbf{m}, \bTheta)] \mathbb{E}[\flin(\bX'; \mathbf{m}, \bTheta)]^\top.
\end{split}
\end{align}
We already know that  $\mathbb{E}[\flin(\bX ; \bTheta)] = f(\bX ; \mathbf{m})$, so we only need to find $\mathbb{E}[ \flin(\bX ; \bTheta) \flin(\bX'; \bTheta)^\top ]$:
\begin{align}
    \begin{split}
    \mathbb{E}_{g_{\bTheta}}&[ \flin(\bX ; \mathbf{m}, \bTheta) \flin(\bX'; \mathbf{m}, \bTheta)^\top ]
    \\
    =
    &\mathbb{E}_{g_{\bTheta}}[ (f(\bX ; \mathbf{m}) + \jac(\bX ; \mathbf{m})(\bTheta - \mathbf{m})) (f(\bX'; \mathbf{m}) + \jac(\bX' ; \mathbf{m})(\bTheta - \mathbf{m}))^\top ]
    \end{split}
    \\
    \begin{split}
    =
    &\mathbb{E}_{g_{\bTheta}}[ f(\bX ; \mathbf{m}) f(\bX' ; \mathbf{m})^\top 
    + 
    (\jac(\bX ; \mathbf{m})(\bTheta - \mathbf{m})) (\jac(\bX' ; \mathbf{m})(\bTheta - \mathbf{m}))^\top
    \\
    &\qquad \qquad + f(\bX ; \mathbf{m}) (\jac(\bX' ; \mathbf{m})(\bTheta - \mathbf{m}))^\top
    + \jac(\bX ; \mathbf{m})(\bTheta - \mathbf{m}) f(\bX' ; \mathbf{m})^\top ]
    \end{split}
    \\
    \begin{split}
    =
    &\mathbb{E}_{g_{\bTheta}}[ f(\bX ; \mathbf{m}) f(\bX' ; \mathbf{m})^\top 
    + 
    \jac(\bX ; \mathbf{m}) (\bTheta - \mathbf{m}) (\bTheta - \mathbf{m})^\top \jac(\bX' ; \mathbf{m})^\top
    \\
    &\qquad \qquad + f(\bX ; \mathbf{m}) (\jac(\bX' ; \mathbf{m})(\bTheta - \mathbf{m}))^\top
    + \jac(\bX ; \mathbf{m})(\bTheta - \mathbf{m}) f(\bX' ; \mathbf{m})^\top ]
    \end{split}
    \\
    \begin{split}
    =
    & f(\bX ; \mathbf{m}) f(\bX' ; \mathbf{m})^\top 
    + 
    \jac(\bX ; \mathbf{m}) \mathbb{E}_{g_{\bTheta}}[ (\bTheta - \mathbf{m}) (\bTheta - \mathbf{m})^\top ] \jac(\bX' ; \mathbf{m})^\top
    \\
    &\qquad \qquad + f(\bX ; \mathbf{m}) (\jac(\bX' ; \mathbf{m})(\underbrace{\mathbb{E}_{g_{\bTheta}}[\bTheta] - \mathbf{m})}_{=0})^\top
    + \jac(\bX ; \mathbf{m})(\underbrace{\mathbb{E}_{g_{\bTheta}}[\bTheta] - \mathbf{m}}_{=0}) f(\bX' ; \mathbf{m})^\top,
    \end{split}
    \end{align}
    where the last line follows from the definition of $g_{\bTheta}$.
    By definition of the covariance, we then obtain
    \begin{align}
    \begin{split}
    &
    \mathbb{E}_{g_{\bTheta}}[ \flin(\bX ; \mathbf{m}, \bTheta) \flin(\bX'; \mathbf{m}, \bTheta)^\top ]
    \\
    &=
    f(\bX ; \mathbf{m}) f(\bX' ; \mathbf{m})^\top 
    + 
    \jac(\bX ; \mathbf{m}) \mathbb{E}_{g_{\bTheta}}[ (\bTheta - \mathbf{m}) (\bTheta - \mathbf{m})^\top ] \jac(\bX' ; \mathbf{m})^\top
    \end{split}
    \\
    &=
     f(\bX ; \mathbf{m}) f(\bX ; \mathbf{m})^\top 
    + 
    \jac(\bX ; \mathbf{m}) \text{Cov}( \bTheta ) \jac(\bX' ; \mathbf{m})^\top.
\end{align}
With this result, we obtain the covariance function
\begin{align}
\begin{split}
    &
    \textrm{Cov}(\flin(\bX ; \mathbf{m}, \bTheta), \flin(\bX'; \mathbf{m}, \bTheta))
    \\
    &=
    \mathbb{E}[ \flin(\bX ; \mathbf{m}, \bTheta) \flin(\bX'; \mathbf{m}, \bTheta)^\top ] - \mathbb{E}[\flin(\bX ; \mathbf{m}, \bTheta)] \mathbb{E}[\flin(\bX'; \mathbf{m}, \bTheta)]^\top
    \end{split}
    \\
    &= \mathbb{E}[ \flin(\bX ; \mathbf{m}, \bTheta) \flin(\bX'; \mathbf{m}, \bTheta)^\top ] - f(\bX ; \mathbf{m}) f(\bX ; \mathbf{m})^\top + \jac(\bX ; \mathbf{m}) \text{Cov}( \bTheta ) \jac(\bX' ; \mathbf{m})^\top
    \\
    &= f(\bX ; \bTheta) f(\bX'; \bTheta)^\top - f(\bX ; \mathbf{m}) f(\bX ; \mathbf{m})^\top + \jac(\bX ; \mathbf{m}) \text{Cov} \bTheta ) \jac(\bX' ; \mathbf{m})^\top
    \\
    &= \jac(\bX ; \mathbf{m}) \Var[ \bTheta ] \jac(\bX' ; \mathbf{m})^\top.
\end{align}
Finally, $\text{Cov} (\bTheta ) = \mathbf{S}$ yields $\textrm{Cov}(\flin(\bX ; \mathbf{m}, \bTheta), \flin(\bX'; \mathbf{m}, \bTheta)) = \jac(\bX ; \mathbf{m}) \mathbf{S} \jac(\bX' ; \mathbf{m})^\top$.
This concludes the proof.
\end{proof}

\clearpage

\begin{customproposition}{2}[Approximate Distribution under Linearized Mapping]
\label{app-prop:linearization_approx}
For a stochastic function $f(\cdot \,; \bTheta)$ defined in terms of stochastic parameters $\bTheta$ distributed according to distribution \mbox{$g_{\bTheta} = \calN(\mathbf{m}, \mathbf{S})$}, denote the linearization of the stochastic function $f(\cdot \,; \bTheta)$ about $\mathbf{m}$ by
\begin{align*}
\label{app-eq:linearization}
    f(\cdot \,; \bTheta) \approx \flin(\cdot \,; \mathbf{m}, \bTheta) \defines f(\cdot \,; \mathbf{m}) + \jac(\cdot \,; \mathbf{m})(\bTheta - \mathbf{m}) ,
    \end{align*}
where $\jac(\cdot \,; \mathbf{m}) \defines (\partial f(\cdot \,; \bTheta) / \partial \bTheta)|_{\bTheta = \mathbf{m}}$ is the Jacobian of $f(\cdot \,; \bTheta)$ evaluated at $\bTheta = \mathbf{m}$.
Then, for a partition of the set of parameters into sets $\alpha$ and $\beta$, a distribution \mbox{$g_{\bTheta} = \calN(\mathbf{m}, \mathbf{S})$} with \mbox{$\bTheta_{\alpha} \perp \bTheta_{\beta}$}, the distribution $\tilde{g}_{\flin(\bX ; \bTheta)}$ can be approximated via the Monte Carlo estimator
\begin{align}
    \hat{\tilde{g}}_{\flin(\bX ; \mathbf{m}, \bTheta)}
    =
    \frac{1}{R} \sum\nolimits_{j=1}^{R} \mathcal{N}\Big(f(\bX ; \mathbf{m}) + \flin_{\alpha}(\bX ; \mathbf{m}, \bTheta_{\alpha})^{(j)}, \jac_{\beta}(\bX ; \mathbf{m}) \mathbf{S}_{\beta} {\jac(\bX' ; \mathbf{m})_{\beta}}^\top \Big) ,
\end{align}
where \mbox{$g_{\bTheta_{\alpha}} = \calN(\mathbf{m}_{\alpha}, \mathbf{S}_{\alpha})$}, $g_{\bTheta_{\beta}} = \calN(\mathbf{m}_{\beta}, \mathbf{S}_{\beta})$, and
\begin{align}
    \flin_{\alpha}(\cdot \,; \mathbf{m}, \bTheta_{\alpha}) \defines \jac_{\alpha}(\cdot \,; \mathbf{m})(\bTheta_{\alpha} - \mathbf{m}_{\alpha}) ,
\end{align}
with $\jac_{\alpha}(\cdot \,; \mathbf{m})$ denoting the columns of the Jacobian matrix corresponding to the sets of parameters $\alpha$ and $\flin_{\alpha}(\bX ; \mathbf{m}, \bTheta_{\alpha})^{(j)}$ for $j=1,...,R$ obtained by sampling parameters from the distribution \mbox{$g_{\bTheta_{\alpha}} =  \calN(\mathbf{m}_{\alpha}, \mathbf{S}_{\alpha})$}.
\end{customproposition}
\begin{proof}
Consider a partition of the set of parameters into sets $\alpha$ and $\beta$ and express the linearized mapping as
\begin{align}
\SwapAboveDisplaySkip
\label{app-eq:linearization_decomposition}
    \flin(\cdot \,; \mathbf{m}, \bTheta)
    =
    \flin_{\alpha}(\cdot \,; \mathbf{m}, \bTheta_{\alpha}) + \flin_{\beta}(\cdot \,; \mathbf{m}, \bTheta_{\beta}) ,
\end{align}
with
\begin{align}
\SwapAboveDisplaySkip
    \flin_{\alpha}(\cdot \,; \mathbf{m}, \bTheta_{\alpha})
    \defines
    \jac_{\alpha}(\cdot \,; \mathbf{m})(\bTheta_{\alpha} - \mathbf{m}_{\alpha}) ,
\end{align}
and
\begin{align}
\SwapAboveDisplaySkip
    \flin_{\beta}(\cdot \,; \mathbf{m}, \bTheta_{\beta})
    \defines
    f(\cdot \,; \mathbf{m}) + \jac_{\beta}(\cdot \,; \mathbf{m})(\bTheta_{\beta} - \mathbf{m}_{\beta}) ,
\end{align}
where $\jac_{\alpha}(\cdot \,; \mathbf{m})$ and $\jac_{\beta}(\cdot \,; \mathbf{m})$ are the columns of the Jacobian matrix corresponding to the sets of parameters $\alpha$ and $\beta$, respectively, and $\bTheta_{\alpha}$ and $\bTheta_{\beta}$ are the corresponding random parameter vectors.

Noting that~\Cref{app-eq:linearization_decomposition} expresses $\flin$ as a sum of (affine transformations of) random variables, we can use the fact that for independent Gaussian random variables $\bX$ and $\mathbf{Y}$, the distribution $h_{\mathbf{Z}}$ of $\mathbf{Z} = \bX + \mathbf{Y}$ is equal to the convolution of the distributions $h_{\bX}$ and $h_{\mathbf{Y}}$ to obtain an approximation to $\flin$.
In particular, for $\mathbf{Z} = \bX + \mathbf{Y}$,
\begin{align}
\SwapAboveDisplaySkip
    f_{\mathbf{Z}}(\mathbf{z})
    =
    \int_{-\infty}^{\infty} f_{\mathbf{Y}}(\mathbf{z} - \mathbf{x}) f_{\mathbf{X}}(\mathbf{x}) \dee \mathbf{x} .
\end{align}
Letting $\mathbf{X} = \flin_{\alpha}(\bX ; \mathbf{m}, \bTheta_{\alpha})$, $\mathbf{Y} = \flin_{\beta}(\bX ; \mathbf{m}, \bTheta_{\beta})$, and $\mathbf{X} = \flin(\bX ; \bTheta)$, we can write
\begin{align}
\begin{split}    
    &
    \tilde{g}_{\flin(\bX ; \mathbf{m}, \bTheta)}(\flin(\bX ; \mathbf{m}, \btheta))
    \\
    &
    =
    \int_{-\infty}^{\infty} \tilde{g}_{\flin_{\beta}(\bX ; \mathbf{m}, \bTheta_{\beta})}(\flin(\bX ; \mathbf{m}, \btheta) - \flin_{\alpha}(\bX ; \mathbf{m}, \btheta_{\alpha})) \tilde{g}_{\flin_{\alpha}(\bX ; \mathbf{m}, \bTheta_{\alpha})}(\flin_{\alpha}(\bX ; \mathbf{m}, \btheta_{\alpha})) \dee \bX ,
    \end{split}
    \\
    &
    =
    \int_{-\infty}^{\infty} \calN(\flin(\bX ; \mathbf{m}, \btheta) \,; \bm{\mu}(\bX, \mathbf{m}, \btheta_{\alpha}, \flin_{\alpha}) , \bm{\Sigma}(\bX, \mathbf{m}, \btheta_{\beta}, \mathbf{S}_{\beta})) \tilde{g}_{\flin_{\alpha}(\bX ; \mathbf{m}, \bTheta_{\alpha})}(\flin_{\alpha}(\bX ; \mathbf{m}, \btheta_{\alpha})) \dee \bX ,
\end{align}
with
\begin{align}
\SwapAboveDisplaySkip
    \bm{\mu}(\bX ; \mathbf{m}, \btheta_{\alpha}, \flin_{\alpha})
    =
    f(\bX ; \mathbf{m}) + \flin_{\alpha}(\bX ; \mathbf{m}, \btheta_{\alpha})
\end{align}
and
\begin{align}
\SwapAboveDisplaySkip
    \bm{\Sigma}(\bX ; \mathbf{m}, \mathbf{S}_{\beta}, \jac_{\beta})
    =
    \jac_{\beta}(\bX ; \mathbf{m}) \mathbf{S}_{\beta} {\jac_{\beta}(\bX ; \mathbf{m})}^\top ,
\end{align}
where we have used the fact that for a Gaussian distribution with mean $m$ and covariance $S$, \mbox{$\calN(z - y ; m, S) = \calN(z ; m + y, S)$}.
We can then approximate the probability density function $\tilde{g}_{\flin(\bX ; \mathbf{m}, \bTheta)}(\flin(\bX ; \btheta))$ via the Monte Carlo estimator
\begin{align}
\begin{split}
    &
    \hat{\tilde{g}}_{\flin(\bX ; \mathbf{m}, \bTheta)}(\flin(\bX ; \mathbf{m}, \btheta))
    \\
    &
    =
    \frac{1}{R} \sum\nolimits_{j=1}^{R} \calN(\flin(\bX ; \mathbf{m}, \btheta) \,; \bm{\mu}(\bX, \mathbf{m}, \flin_{\alpha}(\bX ; \mathbf{m}, \btheta_{\alpha})^{(j)}) , \bm{\Sigma}(\bX ; \mathbf{m}, \mathbf{S}_{\beta}, \jac_{\beta}) )
\end{split}
\end{align}
with $\flin_{\alpha}(\bX ; \mathbf{m}, \btheta_{\alpha})^{(j)} \sim \tilde{g}_{\flin_{\alpha}(\bX ; \mathbf{m}, \bTheta_{\alpha})}$.
Finally, we can express the distribution $\hat{\tilde{g}}_{\flin(\bX ; \mathbf{m}, \bTheta)}$ as
\begin{align}
    \hat{\tilde{g}}_{\flin(\bX ; \mathbf{m}, \bTheta)}
    =
    \frac{1}{R} \sum\nolimits_{j=1}^{R} \mathcal{N}\Big(f(\bX ; \mathbf{m}) + \flin_{\alpha}(\bX ; \mathbf{m}, \bTheta_{\alpha})^{(j)}, \jac_{\beta}(\bX ; \mathbf{m}) \mathbf{S}_{\beta} {\jac_{\beta}(\bX ; \mathbf{m})}^\top \Big) ,
\end{align}
where $g_{\bTheta_{\beta}} = \calN(\mathbf{m}_{\beta}, \mathbf{S}_{\beta})$ and samples $\flin_{\alpha}(\bX ; \mathbf{m}, \bTheta_{\alpha})^{(j)}$ are obtained by sampling parameters from the distribution \mbox{$g_{\bTheta_{\alpha}} =  \calN(\mathbf{m}_{\alpha}, \mathbf{S}_{\alpha})$}.
This concludes the proof.
\end{proof}

\clearpage

\section{Further Empirical Results}
\label{appsec:empirical_results}

\subsection{Tabular Results for Diabetic Retinopathy Diagnosis Tasks}
\label{appsec:retina_tabular}

The results below were reproduced from~\citet{Band2021benchmarking} using the \textsc{retina} benchmark.

\begin{table*}[htb!]
\centering
\caption{
    \textbf{Country Shift.}
    Prediction and uncertainty quality of baseline methods in terms of the area under the receiver operating characteristic curve (AUC) and classification accuracy, as a function of the proportion of data referred to a medical expert.
    All methods are tuned on in-domain validation AUC, and ensembles have $K = 3$ constituent models (true for all subsequent tables unless specified otherwise).
    On in-domain data, \mcd performs best across all thresholds. 
    On distributionally shifted data, no method consistently performs best.
}
\vspace*{-5pt}
\resizebox{1.0\linewidth}{!}{%
\begin{tabular}{@{\extracolsep{2pt}}lcccccc@{}}
\midrule
\midrule
& \multicolumn{2}{c}{No Referral} & \multicolumn{2}{c}{$50\%$ Data Referred} & \multicolumn{2}{c}{$70\%$ Data Referred} \\
\cline{2-3}
\cline{4-5}
\cline{6-7}\\
\textbf{Method}         &
\textbf{AUC (\%) $\uparrow$}            &
\textbf{Accuracy (\%) $\uparrow$}       &
\textbf{AUC (\%) $\uparrow$}            &
\textbf{Accuracy (\%) $\uparrow$}       &
\textbf{AUC (\%) $\uparrow$}            &
\textbf{Accuracy $\uparrow$}       \\
\midrule
\multicolumn{7}{c}{EyePACS Dataset (In-Domain)}\\
\midrule
\map (Deterministic)	& $87.4\pms{1.3}$ & $88.6\pms{0.7}$ & $91.1\pms{1.8}$ & $95.9\pms{0.4}$ & $94.9\pms{1.1}$ & $96.5\pms{0.3}$ \\
\mfvi	& $83.3\pms{0.2}$ & $85.7\pms{0.1}$ & $85.5\pms{0.7}$ & $94.5\pms{0.1}$ & $88.2\pms{0.7}$ & $95.9\pms{0.1}$ \\
\textsc{radial}-\mfvi	& $83.2\pms{0.5}$ & $74.2\pms{5.0}$ & $88.9\pms{0.9}$ & $81.8\pms{6.0}$ & $91.2\pms{1.3}$ & $83.8\pms{5.5}$ \\
\fsvi	& $88.5\pms{0.1}$ & $89.8\pms{0.0}$ & $91.0\pms{0.4}$ & $96.4\pms{0.0}$ & $94.3\pms{0.3}$ & $97.2\pms{0.1}$ \\
\mcd	& $91.4\pms{0.2}$ & $90.9\pms{0.1}$ & $95.3\pms{0.2}$ & $97.4\pms{0.1}$ & $97.4\pms{0.1}$ & $98.1\pms{0.0}$ \\
\textsc{rank}-1	& $85.6\pms{1.4}$ & $87.7\pms{0.8}$ & $87.1\pms{2.3}$ & $95.3\pms{0.5}$ & $90.9\pms{2.0}$ & $96.4\pms{0.4}$ \\
\textsc{deep ensemble}	& $90.3\pms{0.2}$ & $90.3\pms{0.3}$ & $91.7\pms{0.6}$ & $97.2\pms{0.0}$ & $95.0\pms{0.5}$ & $97.9\pms{0.0}$ \\
\mfvi \textsc{ensemble}	& $85.4\pms{0.0}$ & $87.8\pms{0.0}$ & $86.3\pms{0.4}$ & $95.4\pms{0.0}$ & $89.2\pms{0.4}$ & $96.7\pms{0.1}$ \\
\textsc{radial}-\mfvi \textsc{ensemble}	& $84.9\pms{0.1}$ & $74.2\pms{1.5}$ & $91.4\pms{0.2}$ & $83.4\pms{1.7}$ & $93.3\pms{0.3}$ & $85.9\pms{1.6}$ \\
\fsvi \textsc{ensemble}	& $90.3\pms{0.1}$ & $90.6\pms{0.0}$ & $92.1\pms{0.2}$ & $97.1\pms{0.0}$ & $95.2\pms{0.2}$ & $97.8\pms{0.1}$ \\
\mcd \textsc{ensemble}	& $\mathbf{92.5\pms{0.0}}$ & $\mathbf{91.6\pms{0.0}}$ & $\mathbf{95.8\pms{0.1}}$ & $\mathbf{97.8\pms{0.0}}$ & $\mathbf{97.7\pms{0.1}}$ & $\mathbf{98.4\pms{0.0}}$ \\
\textsc{rank}-1 \textsc{ensemble}	& $89.5\pms{0.8}$ & $89.3\pms{0.4}$ & $88.5\pms{1.3}$ & $96.9\pms{0.3}$ & $91.6\pms{1.2}$ & $97.6\pms{0.3}$ \\
\midrule
\multicolumn{7}{c}{APTOS 2019 Dataset (Population Shift)}\\
\midrule

\map (Deterministic)	& $92.2\pms{0.2}$ & $86.2\pms{0.6}$ & $80.1\pms{3.6}$ & $87.6\pms{1.5}$ & $55.4\pms{4.3}$ & $85.4\pms{1.2}$ \\
\mfvi	& $91.4\pms{0.2}$ & $84.1\pms{0.3}$ & $93.8\pms{0.4}$ & $92.1\pms{0.5}$ & $93.0\pms{0.6}$ & $92.7\pms{0.5}$ \\
\textsc{radial}-\mfvi	& $90.7\pms{0.7}$ & $71.8\pms{4.6}$ & $82.0\pms{2.5}$ & $81.5\pms{2.7}$ & $66.4\pms{2.1}$ & $85.9\pms{1.0}$ \\
\fsvi	& $94.1\pms{0.1}$ & $87.6\pms{0.5}$ & $90.6\pms{0.9}$ & $90.7\pms{0.7}$ & $77.2\pms{4.6}$ & $89.8\pms{0.3}$ \\
\mcd	& $94.0\pms{0.2}$ & $86.8\pms{0.2}$ & $87.4\pms{0.3}$ & $88.1\pms{0.2}$ & $65.3\pms{1.7}$ & $88.2\pms{0.4}$ \\
\textsc{rank}-1	& $92.5\pms{0.3}$ & $86.2\pms{0.5}$ & $90.1\pms{2.5}$ & $91.4\pms{1.1}$ & $75.1\pms{7.8}$ & $89.5\pms{1.5}$ \\
\textsc{deep ensemble}	& $94.2\pms{0.2}$ & $87.5\pms{0.1}$ & $91.2\pms{1.9}$ & $92.4\pms{0.9}$ & $67.4\pms{7.3}$ & $90.1\pms{1.2}$ \\
\mfvi \textsc{ensemble}	& $93.2\pms{0.1}$ & $87.0\pms{0.2}$ & $\mathbf{94.9\pms{0.3}}$ & $\mathbf{93.7\pms{0.3}}$ & $\mathbf{94.2\pms{0.3}}$ & $\mathbf{94.0\pms{0.3}}$ \\
\textsc{radial}-\mfvi \textsc{ensemble}	& $91.8\pms{0.2}$ & $69.0\pms{1.9}$ & $78.6\pms{0.6}$ & $79.8\pms{0.9}$ & $60.9\pms{0.3}$ & $86.7\pms{0.2}$ \\
\fsvi \textsc{ensemble}	& $\mathbf{94.6\pms{0.1}}$ & $\mathbf{88.9\pms{0.2}}$ & $90.7\pms{0.5}$ & $91.1\pms{0.6}$ & $74.1\pms{3.4}$ & $89.8\pms{0.2}$ \\
\mcd \textsc{ensemble}	& $94.1\pms{0.1}$ & $87.6\pms{0.1}$ & $86.8\pms{0.2}$ & $88.0\pms{0.2}$ & $62.3\pms{0.4}$ & $87.7\pms{0.2}$ \\
\textsc{rank}-1 \textsc{ensemble}	& $94.1\pms{0.2}$ & $88.3\pms{0.2}$ & $\mathbf{94.9\pms{0.4}}$ & $93.5\pms{0.3}$ & $92.4\pms{1.5}$ & $93.8\pms{0.3}$ \\
\midrule
\midrule
\end{tabular}
}
\label{tab:metrics_country}
\end{table*}

\subsection{UCI Regression}
\label{appsec:uci}

\setlength{\tabcolsep}{11pt}
\begin{table}[htb!]
\vspace*{-10pt}
\small
\centering
\caption{
    This table compares the predictive performance between the method proposed in this paper and the method proposed by~\citet{sun2019fbnn} on six datasets from the UCI database.
    We followed the same training protocol as~\citet{sun2019fbnn} and used the code provided by the authors to load and process the data.
    The same network architecture was used (one hidden layer with 50 hidden units).
    We report the results for the best set of hyperparameters, computed over ten random seeds.
    Lower RMSE and higher log-likelihood are better.
    Best results are shaded in gray.
    The first five rows are small-scale UCI experiments, and the sixth row (``Protein'') is a larger-scale experiment (45,740 data points).
}
\begin{tabular}{lllll}
\midrule
\midrule
         & \multicolumn{2}{c}{RMSE}                                      & \multicolumn{2}{c}{Log-Likelihood}                            \\
         & \multicolumn{1}{c}{\citet{sun2019fbnn}} & \multicolumn{1}{c}{\textbf{Ours}} & \multicolumn{1}{c}{\citet{sun2019fbnn}} & \multicolumn{1}{c}{\textbf{Ours}} \\
\midrule
Boston   & \cellcolor[gray]{0.9} \hspace*{-5pt}$\mathbf{2.378 \pm 0.104}$                & \hspace*{5pt}$3.632 \pm 0.515$                  & \hspace*{-5pt}\cellcolor[gray]{0.9} $\mathbf{-2.301 \pm 0.038}$               & \hspace*{5pt}$-3.150 \pm 0.495$                 \\
Concrete & $4.935 \pm 0.180$                    & \cellcolor[gray]{0.9} $\mathbf{4.177 \pm 0.443}$              & $-3.096 \pm 0.016$                   & \cellcolor[gray]{0.9} $\mathbf{-2.855 \pm 0.116}$             \\
Energy   & $0.412 \pm 0.017$                    & \cellcolor[gray]{0.9} $\mathbf{0.409 \pm 0.060}$              & $-0.684 \pm 0.020$                   & \cellcolor[gray]{0.9}\hspace*{2.5pt}$\mathbf{-0.539 \pm 0.138}$             \\
Wine     & $0.673 \pm 0.014$                    & \cellcolor[gray]{0.9} $\mathbf{0.615 \pm 0.033}$              & $-1.040 \pm 0.013$                   & \cellcolor[gray]{0.9} $\mathbf{-0.959 \pm 0.034}$             \\
Yacht    & $0.607 \pm 0.068$                    & \cellcolor[gray]{0.9} $\mathbf{0.514 \pm 0.242}$              & $-1.033 \pm 0.033$                   & \cellcolor[gray]{0.9} $\mathbf{-0.888 \pm 0.334}$             \\
Protein  & $4.326 \pm 0.019$                    & \cellcolor[gray]{0.9} $\mathbf{4.248 \pm 0.043}$              & $-2.892 \pm 0.004$                   & \cellcolor[gray]{0.9} $\mathbf{-2.866 \pm 0.009}$             \\
\midrule
\midrule
\end{tabular}
\vspace*{-20pt}
\end{table}

\clearpage

\section{Illustrative Examples}
\label{appsec:illustrative_examples}

\subsection{Two Moons Classification Task}

\begin{figure}[h!]
\vspace*{-10pt}
\centering
    \subfloat[\fsvi: Posterior Predictive Mean]{\label{fig:two_moons_fsvi_exp}%
      \includegraphics[width=0.47\linewidth, keepaspectratio]{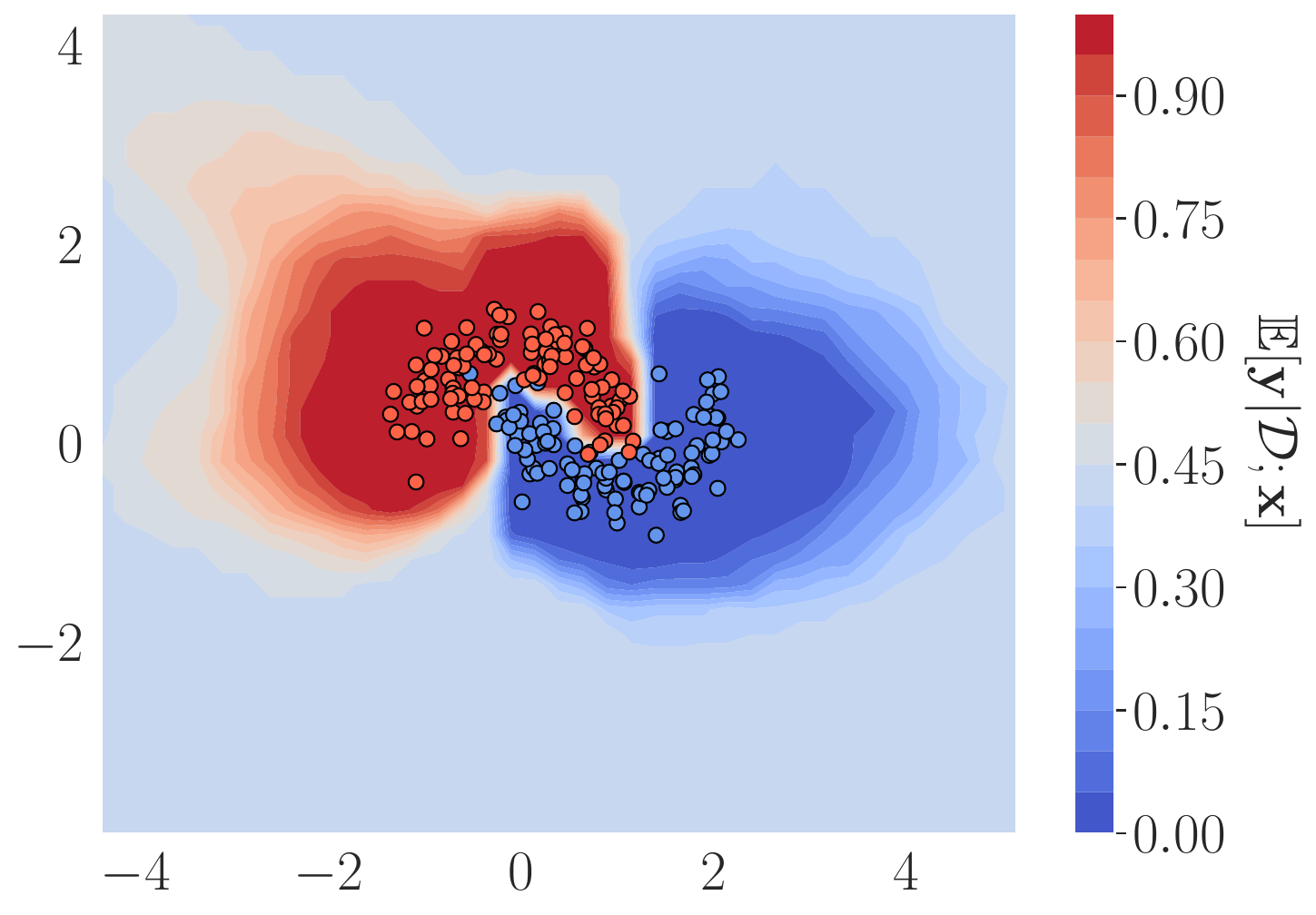}}%
    \hfill
    \subfloat[\fsvi: Posterior Predictive Variance]{\label{fig:two_moons_fsvi_var}%
      \includegraphics[width=0.47\linewidth, keepaspectratio]{figures/icml2021/two_moons_fsvi_predictive_variance.pdf}}
     \\
    \subfloat[\mfvi: Posterior Predictive Mean]{\label{fig:two_moons_mfvi_exp}%
      \includegraphics[width=0.47\linewidth, keepaspectratio]{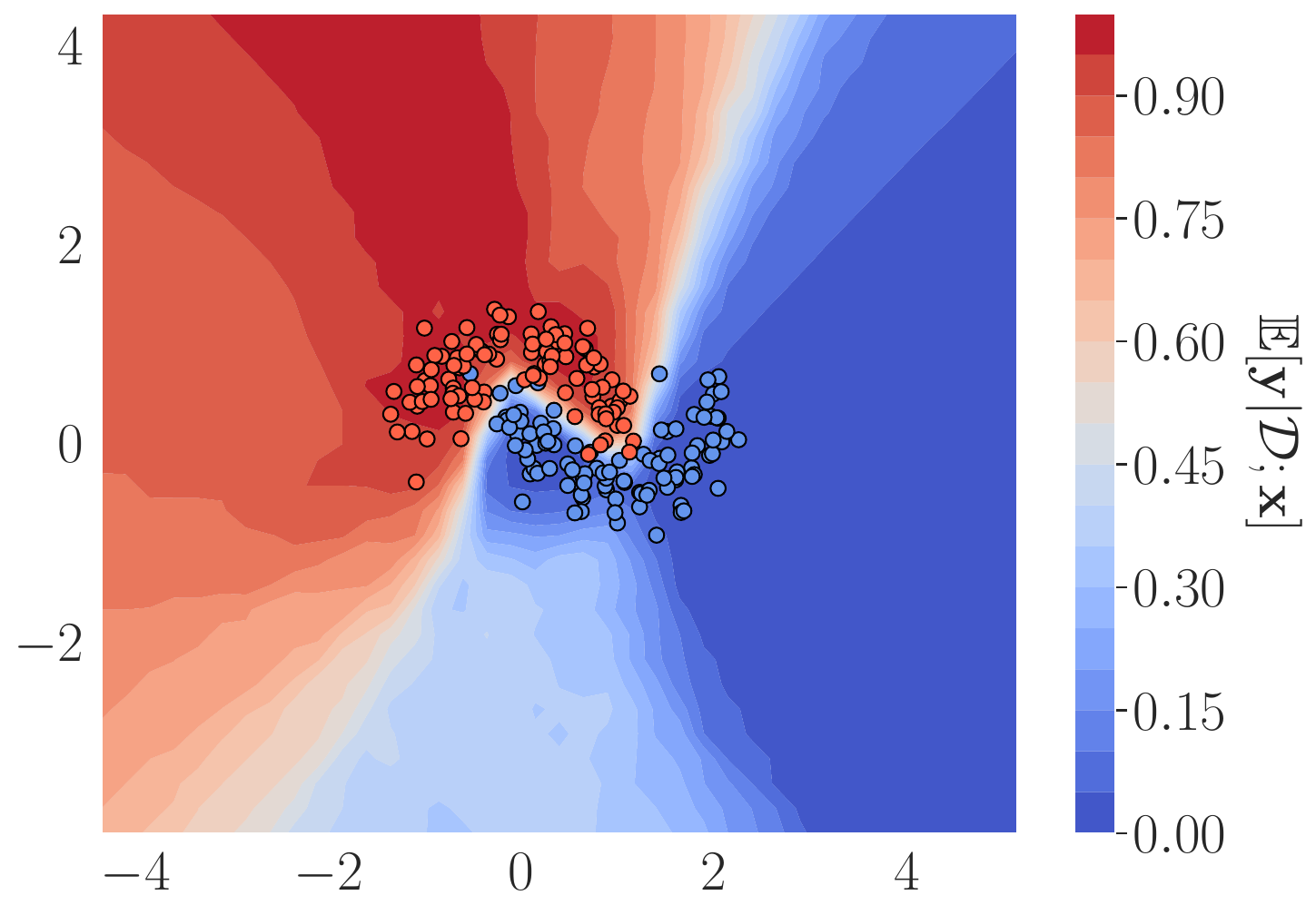}}%
    \hfill
    \subfloat[\mfvi: Posterior Predictive Variance]{\label{fig:two_moons_mfvi_var}%
      \includegraphics[width=0.47\linewidth, keepaspectratio]{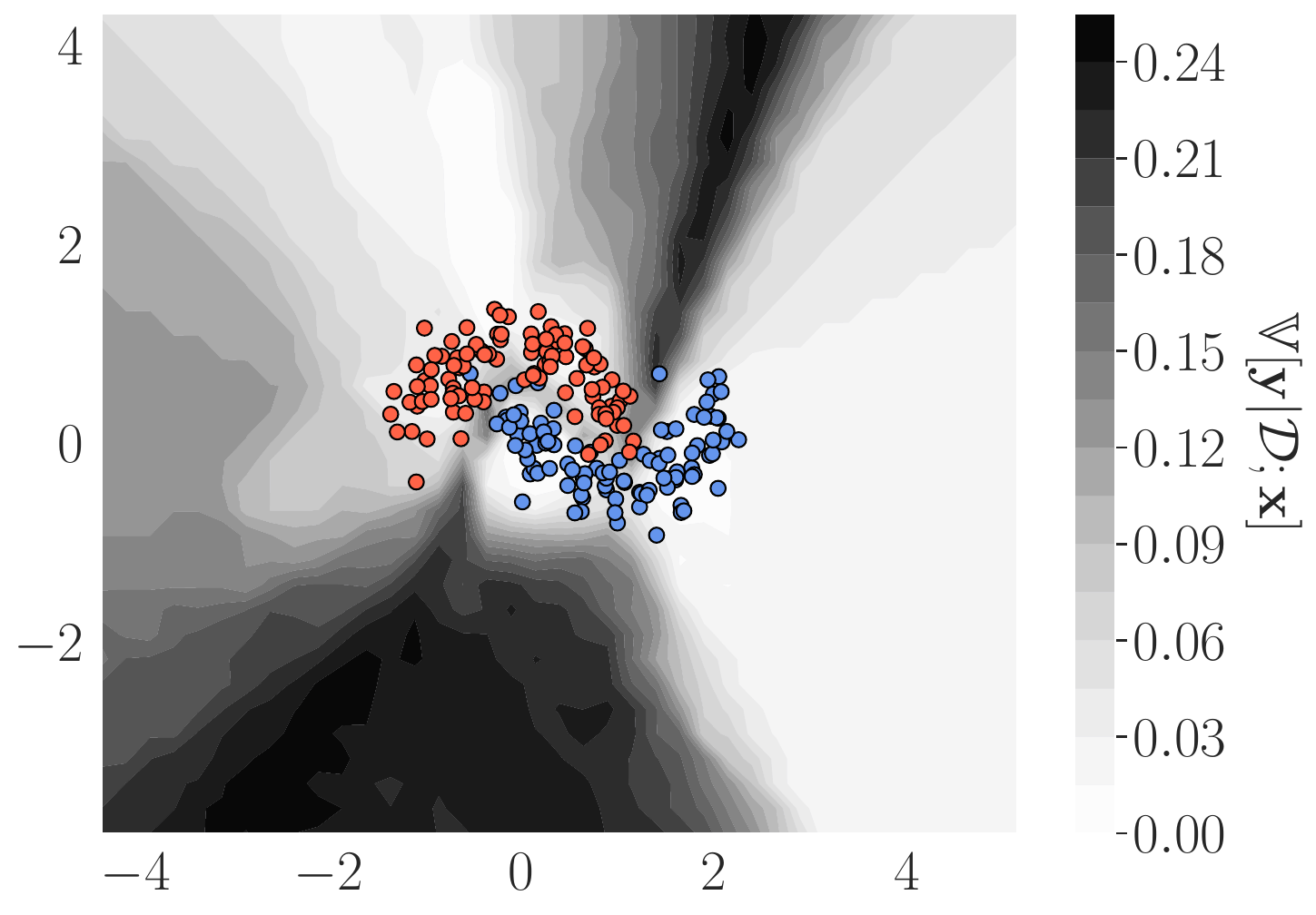}}
     \\
    \subfloat[\textsc{map} Ensemble: Predictive Mean]{\label{fig:two_moons_ensemble_exp}%
      \includegraphics[width=0.47\linewidth, keepaspectratio]{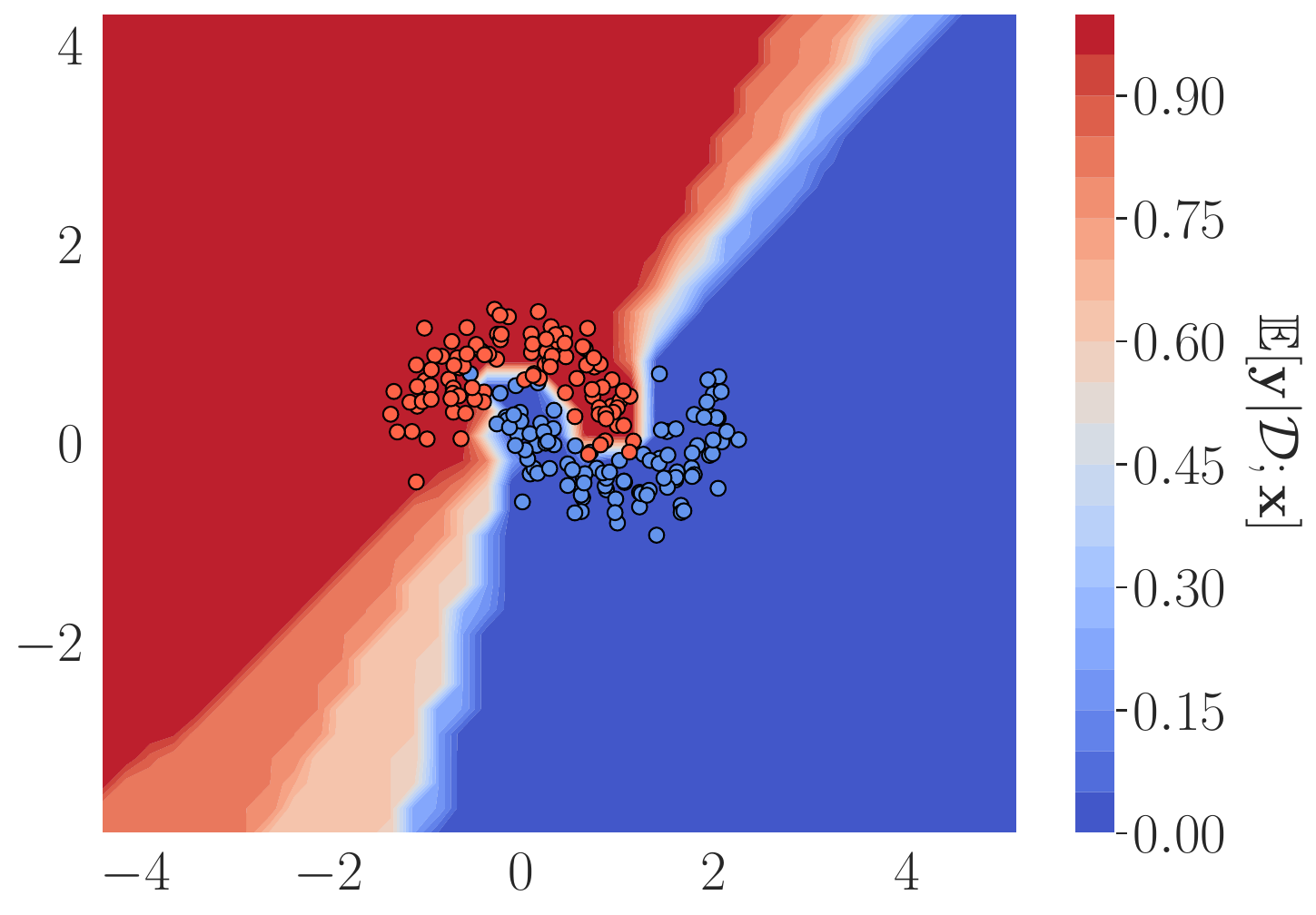}}%
    \hfill
    \subfloat[\textsc{map} Ensemble: Predictive Variance]{\label{fig:two_moons_ensemble_var}%
      \includegraphics[width=0.47\linewidth, keepaspectratio]{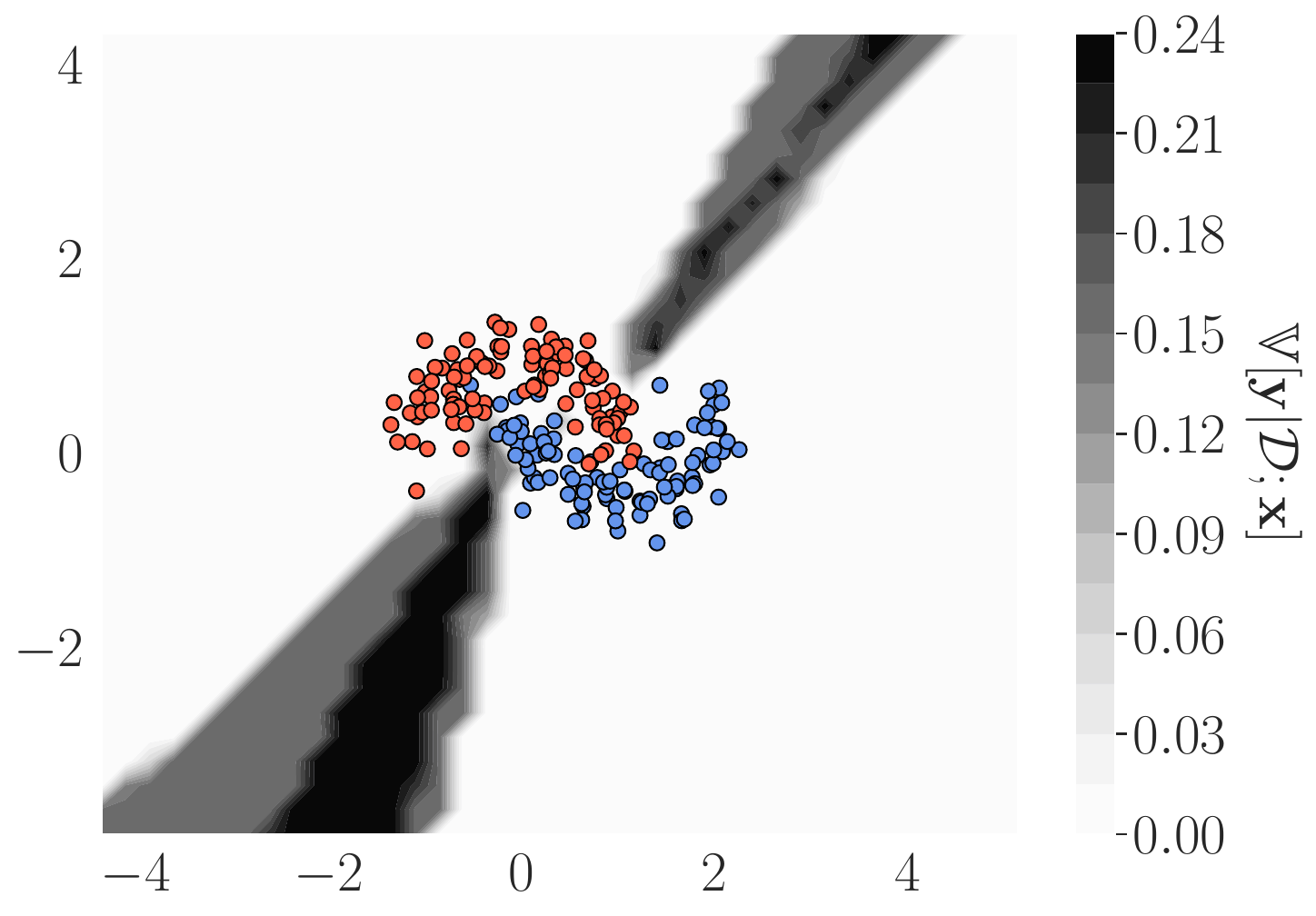}}%
  \label{fig:two_moons}
  \caption{
  Binary classification on the \textit{Two Moons} dataset.
  The plots show the posterior predictive mean and variance of a \bnn trained via \fsvi (\Cref{fig:two_moons_fsvi_exp} and \Cref{fig:two_moons_fsvi_var}), of a \bnn trained via \mfvi (\Cref{fig:two_moons_mfvi_exp} and \Cref{fig:two_moons_mfvi_var}), and an ensemble of \textsc{map} models (\Cref{fig:two_moons_ensemble_exp} and \Cref{fig:two_moons_ensemble_var}).
  The predictive means represent the expected class probabilities and the predictive variance the model's epistemic uncertainty over the class probabilities.
  With \fsvi, the predictive distribution is able to faithfully capture the geometry of the data manifold and exhibits high uncertainty over the class probabilities in areas of the data space of which the data is not informative.
  In contrast, neither \mfvi, nor \textsc{map} ensembles are unable to accurately capture the geometry of the data manifold only exhibit high uncertainty around the decision boundary.
  }
  \vspace*{-20pt}
\end{figure}

\clearpage

\subsection{Synthetic 1D Regression Datasets}

\begin{figure}[h!]
\centering
    \hspace*{-15pt}\subfloat[``Snelson'' Dataset (\citet{SNelsonzoubin_inducing})]{\begin{tabular}[b]{c}%
    \label{fig:snelson_fsvi}%
      \includegraphics[width=0.48\linewidth, keepaspectratio]{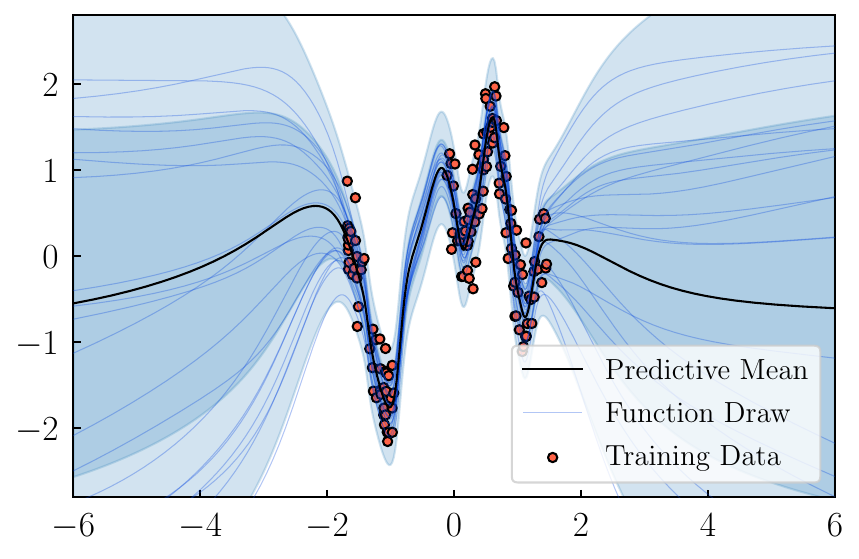}
      \includegraphics[width=0.48\linewidth, keepaspectratio]{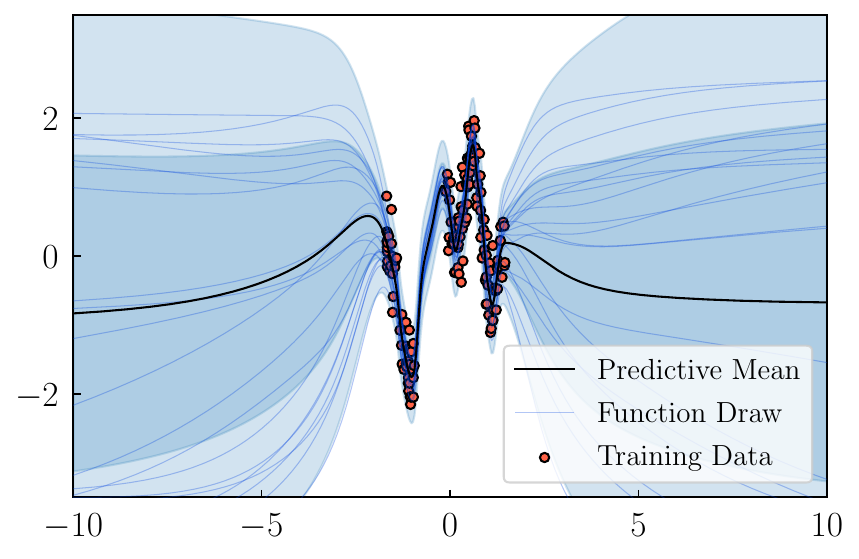}
    \end{tabular}}%
    \\
    \hspace*{-15pt}\subfloat[``OAT-1D'' Dataset (\citet{amersfoort2021variational})]{\begin{tabular}[b]{c}%
    \label{fig:oat1d_fsvi}%
      \includegraphics[width=0.48\linewidth, keepaspectratio]{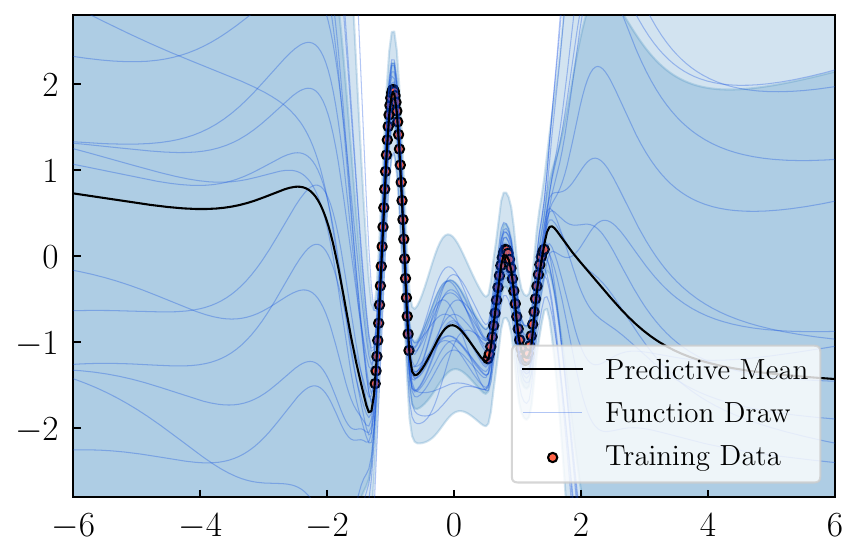}
      \includegraphics[width=0.48\linewidth, keepaspectratio]{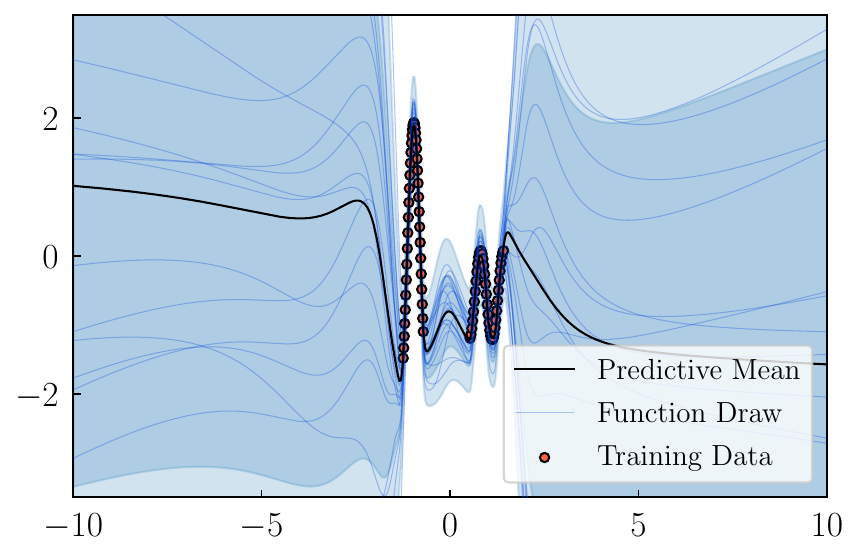}
    \end{tabular}}%
    \\
    \hspace*{-15pt}\subfloat[``Subspace Inference'' Dataset (\citet{izmailov2020subspace})]{\begin{tabular}[b]{c}%
    \label{fig:subspace_inference_fsvi}%
      \includegraphics[width=0.48\linewidth, keepaspectratio]{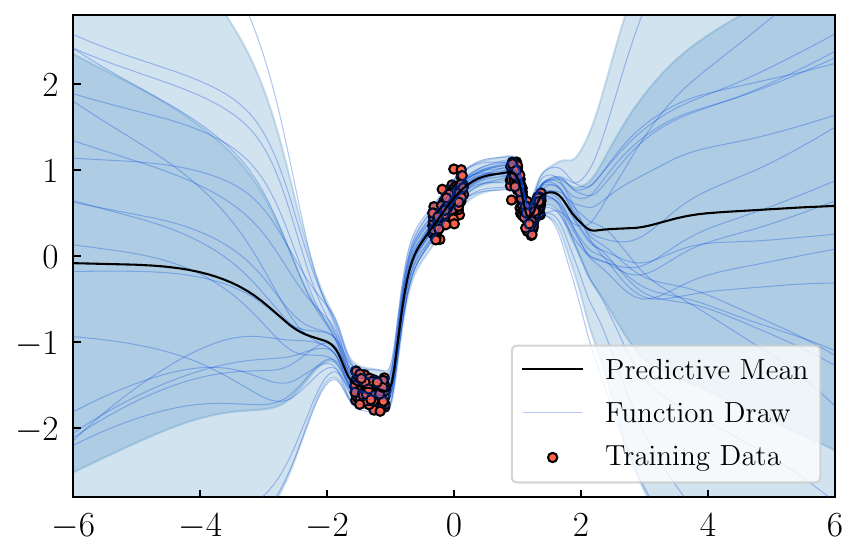}
      \includegraphics[width=0.48\linewidth, keepaspectratio]{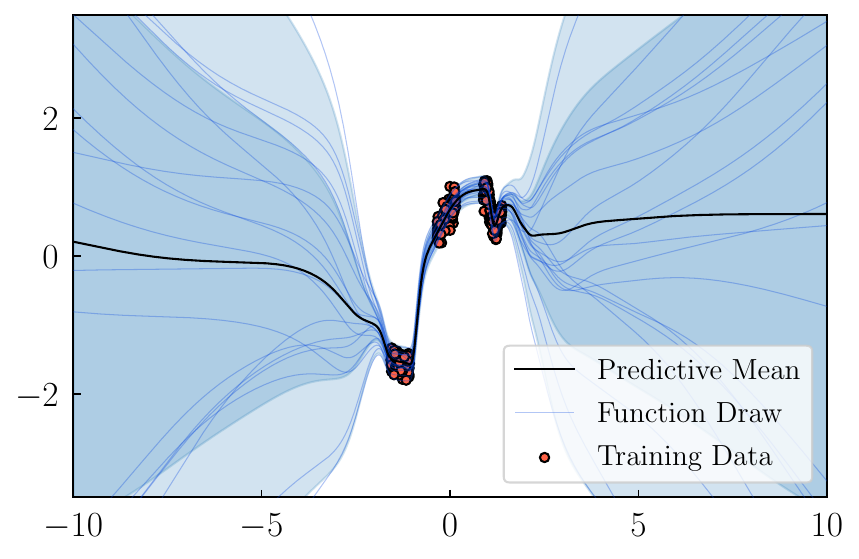}
    \end{tabular}}
  \label{fig:1d_regression}
  \caption{
      1D Regression with \fsvi on a selection of datasets used to demonstrate desirable predictive uncertainty estimates in prior works.
      The left column is zoomed in.
  }
\end{figure}

\clearpage

\section{Implementation, Training, and Evaluation Details}
\label{appsec:model_details}

\subsection{Hyperparameter Selection Protocol}

For \fsvi, we used a holdout validation set (10\% of the training set) to conduct a hyperparameter search over the prior variance, the number of context points used to evaluate the \kld, the context distribution, and the number of Monte Carlo samples used to evaluate the expected log-likelihood.
We selected the set of hyperparameters that yielded the highest validation log-likelihood for all experiments.
We state the hyperparameters selected for the different datasets below.

For other methods, we used a holdout validation set of the same size and selected the best-performing hyperparameters.
We used implementations provided by the authors of \mfvi (radial) and \swag.
All other methods were implemented from scratch unless stated otherwise.

\subsection{FashionMNIST vs. MNIST/NotMNIST}

We train all model on the FashionMNIST dataset and evaluate the models' predictive uncertainty performance on out-of-distribution data on the MNIST dataset.
Both datasets consist of images of size $28 \times 28$ pixels.
The FashionMNIST dataset is normalized to have zero mean and a standard deviation of one.
The MNIST dataset is normalized with the same transformation, that is, using the same mean and standard deviation used for the in-distribution data.
We chose FashionMNIST/MNIST instead of MNIST/NotMNIST because the latter is notably easier than the former.

In this experiment, a network architecture with two convolutional layers of 32 and 64 $3 \times 3$ filters and a fully-connected final layer of 128 hidden units is used.
A max pooling operation is placed after each convolutional layer and ReLU activations are used.
We do not use batch normalization.
All models are trained for 30 epochs with a mini-batch size of 128 using SGD with a learning rate of $5 \times 10^{-3}$, momentum (with momentum parameter 0.9), and a cosine learning rate schedule with parameter $0.05$.

For \fsvi with $p_{\bX_{\calC}}=$random monochrome, we sampled 50\% of the context points for each gradient step from the mini-batch and the other 50\% according to the method described in~\Cref{appsec:inducing_inputs}.
For \fsvi with $p_{\bX_{\calC}}$= KNIST, we used the KMNIST dataset.

\subsection{CIFAR-10 vs. SVHN}

We train all model on the CIFAR-10 dataset and evaluate the models' predictive uncertainty performance on out-of-distribution data on the SVHN dataset.
Both datasets consist of images of size $32 \times 32 \times 3$, with RBG channels.
The CIFAR-10 dataset is normalized to have zero mean and a standard deviation of one.
The SVHN dataset is normalized with the same transformation, that is, using the same mean and standard deviation used for the in-distribution data.
The training data is augmented with random horizontal flips (with a probability of 0.5) and random crops (4 zero pixels on all sides). 

In this experiment, a standard ResNet-18 network architecture was used.
All models are trained for 200 epochs with a mini-batch size of 128 using SGD with a learning rate of $5 \times 10^{-3}$, momentum (with momentum parameter 0.9), and a cosine learning rate schedule with parameter $0.05$.

For \fsvi with $p_{\bX_{\calC}}=$random monochrome, we sampled 100\% of the context points for each gradient step from the mini-batch and the other 50\% according to the method described in~\Cref{appsec:inducing_inputs}.
For \fsvi with $p_{\bX_{\calC}}$= CIFAR-100, we used the CIFAR-100 dataset.

\subsection{Diabetic Retinopathy Diagnosis}
\label{appsec:retina}

\paragraph{Prediction and Expert Referral.}
In real-world settings where the evaluation data may be sampled from a shifted distribution, incorrect predictions may become increasingly likely.
To account for that possibility, predictive uncertainty estimates can be used to identify datapoints where the likelihood of an incorrect prediction is particularly high and refer them for further review.
We consider a corresponding selective prediction task, where the predictive performance of a given model is evaluated for varying expert referral rates.
That is, for a given referral rate of $\gamma \in [0, 1]$, a model's predictive uncertainty is used to identify the $\gamma$ proportion of images in the evaluation set for which the model's predictions are most uncertain.
Those images are referred to a medical professional for further review, and the model is assessed on its predictions on the remaining $(1 - \gamma)$ proportion of images.
By repeating this process for all possible referral rates and assessing the model's predictive performance on the retained images, we estimate how reliable it would be in a safety-critical downstream task, where predictive uncertainty estimates are used in conjunction with human expertise to avoid harmful predictions.
Importantly, selective prediction tolerates out-of-distribution examples.
For example, even if unfamiliar features appear in certain images, a model with reliable uncertainty estimates will perform better in selective prediction by assigning these images high epistemic (and predictive) uncertainty, therefore referring them to an expert at a lower $\gamma$.

For all methods, experiments are performed using a ResNet-50 network architecture.
Training and evaluation scripts as well as model checkpoints can be found at
\begin{center}
    \href{https://github.com/google/uncertainty-baselines/tree/main/baselines/diabetic_retinopathy_detection}{\texttt{github.com/google/uncertainty-baselines/.../diabetic\_retinopathy\_detection}}.
\end{center}

\vspace*{-5pt}  
\subsection{Two Moons}

In this experiment, we use a multi-layer perceptron (MLP) consisting of two fully-connected layers with 30 hidden units each and tanh activations.
We train all models with a learning rate of $10^{-3}$.

For \fsvi, we sampled context points uniformly from $[-10,10] \times [-10,10]$.

\vspace*{-5pt}
\subsection{1D Regression}

In this experiment, we use a multi-layer perceptron (MLP) consisting of two fully-connected layers with 100 hidden units each and ReLU activations.

For \fsvi, we sampled context points uniformly from $[-10,10]$.

\vspace*{-5pt}
\subsection{Further Implementation Details}

We use the Adam optimizer with default settings of $\beta_1 = 0.9$, $\beta_2 = 0.99$ and $\epsilon = 10^{-8}$ for all experiments.
The deterministic neural networks that were used for the ensemble were trained with a weight decay of $\lambda$ = 1e-1.
\mfvi (tempered) was trained with a KL scaling factor of 0.1 to obtain a cold posterior.

\vspace*{-5pt}
\subsection{Selection of Context Distribution}
\label{appsec:inducing_inputs}

We estimate the supremum at every gradient step by sampling a set of context points $\bX_{\calC}$ from a distribution $p_{\bX_{\calC}}$ at every gradient step.
For tasks with image inputs, we construct a distribution $p_{\bX_{\calC}}$, defined as a uniform distribution over images with monochromatic channels.
\edit{To generate a sample from this ``monochrome images'' distribution, we first take all images in the training data, flatten each channel, and stack the flattened image channels into a single vector each.
We then draw a random element (i.e., a pixel) from each channel vector and then use these pixels to generate a monochrome image of a given resolution by setting every channel equal to the value of the pixel that was drawn.}
For regression tasks with a $D$-dimensional input space, $p_{\bX_{\calC}}$ is defined as a uniform distribution with lower and upper bounds set to the empirical lower and upper bounds of the training data.
For further details
on the effect of different sampling schemes on the posterior predictive distribution's performance,
see~\Cref{appsec:empirical_results}.

\vspace*{-5pt}
\subsection{Compute Resources}
\label{appsec:compute}

All experiments were carried out on an Nvidia V-100 GPU with 32GB of memory.

\vspace*{-30pt}

\clearpage

\end{appendices}

\end{document}